\documentclass[10pt]{article} % For LaTeX2e
\usepackage[preprint]{tmlr}

\usepackage{amsmath,amsfonts,bm}

\def\eqref#1{equation~\ref{#1}}
\def\1{\bm{1}}

\DeclareMathAlphabet{\mathsfit}{\encodingdefault}{\sfdefault}{m}{sl}
\SetMathAlphabet{\mathsfit}{bold}{\encodingdefault}{\sfdefault}{bx}{n}

\usepackage{hyperref}
\usepackage{url}
\usepackage{wrapfig,lipsum,booktabs}
\usepackage{amsmath,amsfonts,amssymb}
\usepackage{amsthm}
\usepackage{graphicx}
\usepackage{color}
\usepackage[normalem]{ulem}
\usepackage{diagbox}
\usepackage{makecell}
\usepackage{multicol}
\usepackage{multirow}
\usepackage{tabularx}
\usepackage{paralist}
\usepackage{array}
\usepackage{setspace}

\usepackage{listings}
\usepackage{color}

\definecolor{dkgreen}{rgb}{0,0.6,0}
\definecolor{gray}{rgb}{0.5,0.5,0.5}
\definecolor{mauve}{rgb}{0.58,0,0.82}

\def\name{\textsc{LRPE}}
\def\lrpe{\textsc{LRPE}}

\newcommand{\ig}{\textit{i}.\textit{e}.}
\newcommand*{\tran}{^{\mkern-1.5mu\mathsf{T}}}

\newcommand*{\her}{^{\mathsf{H}}}

\title{Linearized Relative Positional Encoding}

\author{{$^{2}$Zhen Qin\ \ \ $^{2,3}$Weixuan Sun\ \ \ $^{2}$Kaiyue Lu\ \ \ $^{4}$Hui Deng\ \ \ $^{3}$Dongxu Li\ \ \ $^{2}$Xiaodong Han
}\\
{\textbf{ $^{4}$Yuchao Dai ~ $^{5}$Lingpeng Kong ~ $^{1,2}$Yiran Zhong}\thanks{Indicates the corresponding author. Email: \texttt{zhongyiran@gmail.com}} 
}\\
\addr{ $^{1}$Shanghai AI Laboratory \qquad $^{2}$OpenNLPLab  \qquad $^{3}$Australian National University\\ \ $^{4}$Northwestern Polytechnical University\qquad $^{5}$The University of Hong Kong}
}
\def\month{07}  % Insert correct month for camera-ready version
\def\year{2023} % Insert correct year for camera-ready version
\def\openreview{\url{https://openreview.net/forum?id=xoLyps2qWc}} % Insert correct link to OpenReview for camera-ready version

\begin{document}

\maketitle

\begin{abstract}
Relative positional encoding is widely used in vanilla and linear transformers to represent positional information. However, existing encoding methods of a vanilla transformer are not always directly applicable to a linear transformer, because the latter requires a decomposition of the query and key representations into separate kernel functions. Nevertheless, principles for designing encoding methods suitable for linear transformers remain understudied. In this work, we put together a variety of existing linear relative positional encoding approaches under a canonical form and further propose a family of linear relative positional encoding algorithms via \emph{unitary transformation}. Our formulation leads to a principled framework that can be used to develop new relative positional encoding methods that preserve linear space-time complexity. Equipped with different models, the proposed linearized relative positional encoding (\name) family derives effective encoding for various applications. Experiments show that compared with existing methods, \name~achieves state-of-the-art performance in language modeling, text classification, and image classification. Meanwhile, it emphasizes a general paradigm for designing broadly more relative positional encoding methods that are applicable to linear transformers. The code is available at \href{https://github.com/OpenNLPLab/Lrpe}{https://github.com/OpenNLPLab/Lrpe}.
\end{abstract}

\section{Introduction}
\label{intro}
Transformers have achieved remarkable progress in natural language processing~\citep{devlin-etal-2019-bert,radford2019language,brown2020language}, computer vision~\citep{vit,swintransformer,vivit} and audio processing~\citep{karita19_interspeech,zhang2020transformer,gulati20_interspeech,sun2022locality}.
As an important ingredient in transformers, positional encoding assigns a unique representation for each position of a token in a sequence so that the transformers can break the permutation invariance property.
Among these encoding methods, absolute positional encoding~\citep{vaswani2017attention,sukhbaatar2015end,devlin-etal-2019-bert,liu2020roberta} maps each individual position index into a continuous encoding.
Whereas relative positional encoding~\citep{shaw-etal-2018-self,su2021roformer,horn2021translational,liutkus2021relative,huang-etal-2020-improve,raffel2019exploring} generates encoding for each query-key pair, representing their relative positional offset. 
We focus on relative positional encoding as they are not constrained by input lengths~\citep{chen2021permuteformer} while showing superior performance~\citep{shaw-etal-2018-self}.

Linear transformers~\citep{chen2021permuteformer,zhen2022cosformer,qin-etal-2022-devil,liu2022neural,lu2022linear}
attract more attention recently as they can achieve linear space-time complexity 
with respect to input sequence length while maintaining comparable performance with vanilla transformers.
Most existing linear transformers use absolute positional encoding methods to encode positional information, since most existing relative positional encoding methods are designed for vanilla transformers and are not directly applicable to linear transformers.
The main cause behind this limitation is that linear transformers decompose key and value representations in the self-attention modules into separate kernel functions to achieve linear space-time complexity.
Such an additional requirement on the decomposibility is not always satisfied by existing relative positional encoding methods.
On the other hand, despite some individual works~\citep{zhen2022cosformer,chen2021permuteformer}, general principles for designing relative positional encoding for linear transformers remain largely understudied. A recent work, RoPE~\cite{su2021roformer} proposes a new set of multiplicative encoding solutions based on rotational positional encoding and can be applied to linear transformers. In Appendix~\ref{imp_detail}, we show that RoPE can be seen as a special form of {\name}.

\begin{figure*}[t]
  \begin{center}
  {\includegraphics[width=1\linewidth]{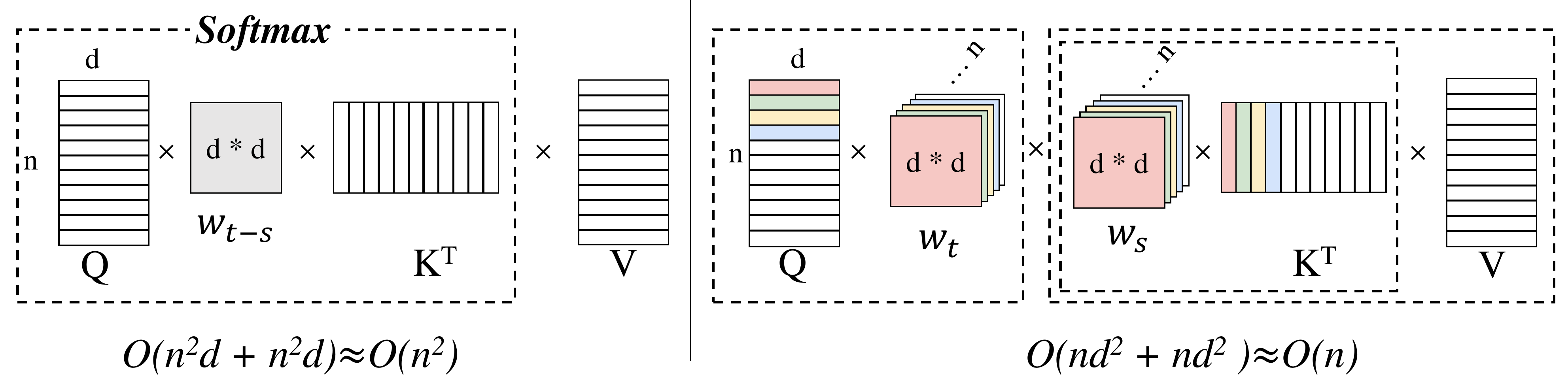}} 
   \vspace{-8mm}
  \end{center}
\caption{Illustration of existing relative positional encoding (left) and the proposed {\name} (right). $\mathbf{Q}$, $\mathbf{K}$, and $\mathbf{V}$ are all in the shape of $n$ by $d$, where $n$ is input length and $d$ is feature dimension.
Tensors in the same dashed line box are associated for computation.
In the vanilla relative positional encoding, query key attention has to be calculated first, leading to a quadratic complexity. $W_{t-s}$ refers to relative positional encoding, where $t,s$ are two positional indices on the query and key, respectively. Our {\name} achieves a decomposable encoding, \ig, $W_t$ and $W_s$ are only dependent on positions of the query and key, making it fully compatible with linear transformers. 
When dealing with long sequences, $d\ll n$, the computation complexity is dominated by $n$, rendering $d$ negligible.
}
  \label{fig:intro}
  \vspace{-4mm}
\end{figure*}

In this work, we aim to bridge this gap and study the principal framework to develop relative positional encoding applicable to linear transformers. 
To this end, we start by presenting a canonical form of relative positional encoding, which reveals that differences in existing encoding methods boil down to choices of a set of query, key, and relative positional matrix \emph{primitives}.
By properly selecting and composing these primitives, we could derive various existing encoding methods for transformers.

Taking advantage of the canonical form, we introduce the main contribution of our work, \ig, a special family of relative positional encoding methods called \emph{linearized relative positional encoding} (\name).
Specifically, we supply a sufficient condition for designing compatible encoding methods, especially for linear transformers, and prove that the linearized relative positional encoding is a unitary transformation.
The benefits of using unitary transformation are twofold.
On one side, since it is derived 
from the decomposable positional matrix, it can maintain the linear space-time complexity as shown in Fig.~\ref{fig:intro}.
Second, the unitary transformation property allows us to effectively derive the family of closed-form solutions.
In particular, we show that a number of encoding methods pertain to the {\name} family, including those used in RoPE~\citep{su2021roformer} and PermuteFormer~\citep{chen2021permuteformer}.

Furthermore, {\name} sheds light on a simple yet flexible theoretical paradigm to develop new effective relative positional encoding.
To demonstrate this, we derive non-exhaustively three additional {\name} encoding methods by parameterizing the generic solution differently, including solutions living in either real or complex domains.
Since unitary transformations are special cases of a relative positional matrix, {\name} is applicable to linear transformers and exclusively suitable within encoder and/or decoder layers.
We experimentally demonstrate the effectiveness of the {\name} family on autoregressive and bidirectional language modeling, text classification, and image classification.
Results show that {\name} achieves superior capability in representing relative positional information, commonly resulting in unrivaled performance than previous encoding methods.

In summary, our main contributions are as follows:
\begin{compactitem}
    \item We present a canonical form of relative positional encoding, which derives most existing relative positional encoding methods as its special case, including those used in linear transformers.
    \item Based on the canonical form, we propose linearized relative position encoding ({\name}), a simple yet principal formulation to derive an encoding \emph{family} that respects the linear space-time complexity in linear transformers.
    We show several existing relative positional encoding methods in linear transformers are in {\name} family. We also provide additional solutions from this generic form. 
    \item Experiments on various downstream tasks, such as language modeling, text classification, and image classification show that the {\name} family is more \emph{robust} and consistently produces better results across tasks than previous relative encoding methods, are \emph{flexible} in being plugged into  encoder and/or decoder layers in linear models.
    In addition, it is \emph{generic} to derive existing and potentially new encoding methods.
\end{compactitem}

\section{Background and Preliminary}
In this section, we provide preliminary knowledge and describe related work
to facilitate the rest discussions.
In the following, we denote the $k$-th row of matrix $\mathbf{M}$ as $\mathbf{m}_k\tran$, the $d$-dimensional identity matrix as $\mathbf {I}_d$. We omit the subscript $d$ when it is unambiguous from the context. The complete list of notations can be found in Appendix \ref{math_notation}. 
\subsection{Transformer and its linearization}

We first briefly review vanilla transformer \citep{vaswani2017attention} and its linearization \citep{katharopoulos2020transformers}.
The key component of transformer models is the self-attention block, which involves three matrices $\mathbf{Q}$ (\textbf{Query}), $\mathbf{K}$ (\textbf{Key}) 
and $\mathbf{V}$(\textbf{Value}); each of them is a linear projection taking $\mathbf X\in \mathbb R^{n\times d}$ as input:
\begin{equation}
\begin{gathered}
    \mathbf{Q} = \mathbf{X}\mathbf{W}_Q  ,
 \mathbf{K} =\mathbf{X}\mathbf{W}_K,
\mathbf{V} =\mathbf{X}\mathbf{W}_V \in \mathbb{R}^{n\times d}.
\end{gathered}
\end{equation}
The output $\mathbf{O}\in \mathbb R^{n\times d}$ is computed using the Softmax weighted sum:
\begin{equation}
\begin{aligned}
\mathbf{O}&= \mathrm{Softmax}(\mathbf{Q} \mathbf{K}\tran / \sqrt{d}) \mathbf{V}.
\end{aligned}
\end{equation}
The computation overhead of the vanilla transformer grows quadratically with respect to the sequence length $n$, which becomes the bottleneck for transformers to handle long input sequences.
\textbf{Linearization} of self-attention aims to reduce the computation complexity to linear~\citep{katharopoulos2020transformers,ke2021rethinking,zhen2022cosformer,vyas2020fast,peng2021random,xiong2021nystromformer,2206.10552}, typically achieved via a decomposable kernel function $\phi: \mathbb R^d\to \mathbb R^{\bar d}$. Specifically, the output of linear attention is computed as:
\begin{equation}
\begin{aligned}
\mathbf{O}&=\mathbf \Delta^{-1} \phi(\mathbf{Q}) [\phi(\mathbf{K})\tran\mathbf V ],\\
\mathbf{\Delta} &= \mathrm{diag}(\phi(\mathbf{Q}) [\phi(\mathbf{K})\tran {\mathbf 1}_n]). 
\end{aligned}
\end{equation}
The key property of linear attention is the \textbf{decomposability} of the kernel function.
This enables to compute $\phi(\mathbf{K})\tran \mathbf{V} \in \mathbb R^{d\times d}$ first, which leads to the $O(nd^2)$ complexity,
further reducing to $O(n)$ with longer inputs ($d\ll n$).
See Appendix \ref{compute} for a detailed discussion.

\subsection{Positional encoding}
Self-attention is capable of parallel sequence processing but cannot capture positional information of each token. To address this issue, positional encoding methods are proposed, which can be generally categorized into two groups: absolute positional encoding and relative positional encoding.

\textbf{Absolute positional encoding} employs handcraft functions~\citep{vaswani2017attention,sukhbaatar2015end}  or learnable encoding lookup tables $\mathbf P\in \mathbb R^{n\times d}$~\citep{devlin-etal-2019-bert,liu2020roberta} to represent position indices as encodings. These encodings are then combined with the context vector additively:
\begin{equation}
\begin{gathered}
\mathbf q_s =\mathbf  W_Q(\mathbf x_s +\mathbf  p_s), 
\mathbf k_s =\mathbf W_K(\mathbf x_s +\mathbf  p_s),\\
\mathbf v_s =\mathbf W_V(\mathbf x_s +\mathbf  p_s),
\end{gathered}
\end{equation}
\noindent
where the encoding formulation only depends on the absolute position index $s$, and the positional encoding size is restricted by the input sequence length.

\textbf{Relative positional encoding} considers relative position offsets between two input tokens \citep{shaw-etal-2018-self,qin2023toeplitz}, \ig,
\begin{equation}
\begin{aligned}
\mathbf e_{st}=\mathbf x_s\tran\mathbf W_Q\tran\mathbf  W_K\mathbf  x_t + f(\mathbf x_s,\mathbf  x_t, t- s),
\end{aligned}
\end{equation}
\noindent
where $s, t$ are the two positional indeices, $\mathbf e_{st}$ denotes the attention score before softmax.
Compared to absolute positional encoding, relative positional encoding generally achieves better performance as it can handle variable input length~\citep{chen2021permuteformer}. However,  extra cost on computation and memory makes it not so efficient than absolute positional encoding~\citep{likhomanenko2021cape}.

Most existing relative positional encoding methods~\citep{raffel2019exploring,shaw-etal-2018-self,huang-etal-2020-improve,chi2022kerple} require computing query-key attention
$\mathbf{Q}\mathbf{K}\tran$ and combine with relative positional information, 
which incurs quadratic complexity.
In contrast, linear attention avoids such a query-key product to achieve the linear complexity.
Therefore, common relative positional encoding methods are usually not applicable in linear transformers.

\section{Our Method}
\label{Method}
In this section, we present our main technical contribution on linearized relative positional encoding, which is an encoding family that preserves linear space-time complexity.
Specifically, we start by presenting a canonical form of relative positional encoding and show that existing encoding methods can be derived by instantiating the canonical form with different choices of so-called primitive queries, keys, and positional matrices in Section \ref{standard}.
When imposing the decomposability constraint on this canonical form, we obtain a sufficient condition for linearized relative positional encoding (LRPE) and derive a family of concrete solutions in real and complex domains in Section~\ref{linear}. We provide an implementation sketch in Section~\ref{imple}. 
\subsection{Canonical form of relative positional encoding}
\label{standard}
In order to better establish connections between existing relative positional encoding methods and understand their design principles, we first present a canonical form of relative positional encoding in this section.
In particular, given a query $\mathbf{q}_s$ and key $\mathbf{k}_s$ pair, their relative positional encoding $f_{\mathrm{rel}}:\mathbb C^d \times \mathbb C^d \to \mathbb C$ can be represented as:
\begin{equation}
f_{\mathrm{rel}}(\mathbf{q}_s, \mathbf{k}_t)= \sum_{l=1}^m  ({\mathbf{\hat q}_s^{(l)}}){\her} \mathbf{W}_{t-s}^{(l)} \mathbf{\hat k}_t^{(l)} ,\label{eq:rel}
\end{equation}
where $\mathsf{H}$ represents \textbf{conjugate transposition} and $m$ represents number of primitives. We refer $\mathbf{\hat q}_s^{(l)}\in \mathbb C^{d^{(l)}_1}, \mathbf{\hat k}_t^{(l)}\in \mathbb C^{ d^{(l)}_2 }, \mathbf{W}_{t-s}^{(l)}\in \mathbb C^{d^{(l)}_1 \times d^{(l)}_2}$ as query, key and relative positional matrix \textbf{\emph{primitives}}, respectively, used as constituent components to construct the relative positional encoding. Note that query primitives do not always indicate a reliance on query embeddings, similarly for other primitives.
For example, an identify matrix can also serve as a primitive, as we will show shortly in Section~\ref{sec:te}.

To demonstrate Eq.~\ref{eq:rel} is a generic formulation, we show that it flexibly induces a wide range of existing relative encoding methods~\citep{shaw-etal-2018-self,su2021roformer,horn2021translational,liutkus2021relative,huang-etal-2020-improve,raffel2019exploring} by selecting and compositing different choices of primitives.
Among them, we highlight four examples in the following section and leave the complete discussions in the Appendix \ref{srpe}.

\subsubsection{Typical encoding examples}\label{sec:te}
\textbf{Additive.}~~In~\citep{huang-etal-2020-improve}, the relative positional encoding is formulated as an extra additive term to the query-key inner-product:
\begin{equation}
f_{\mathrm{rel}}(\mathbf{q}_s, \mathbf{k}_t)= \mathbf{q}_s\her \mathbf{k}_t + w_{t-s},
\end{equation}
% In this situation, we have
\noindent
which can be derived by including an extra identity term as a primitive, formally denoted as:
\begin{equation}
\begin{aligned}
m&=2, \\
\mathbf{\hat q}_s^{(1)}= \mathbf{q}_s,
\mathbf{\hat k}_t^{(1)}&=\mathbf{k}_t,
\mathbf{W}_{t-s}^{(1)}=\mathbf{I}_d,\\
\mathbf{\hat q}_s^{(2)}=\mathbf{I}_d,\mathbf{\hat k}_t^{(2)}&=\mathbf{I}_d,
\mathbf{W}_{t-s}^{(2)}=w_{t-s} \mathbf{I}_d.
\end{aligned}
\end{equation}
\textbf{Multiplicative.} In RoPE~\citep{su2021roformer}, the relative positional encoding works in the form of the weighted inner product:
\begin{equation}
f_{\mathrm{rel}}(\mathbf{q}_s, \mathbf{k}_t)= \mathbf{q}_s\her \mathbf {W}_{t-s} \mathbf{k}_t, 
\end{equation}
which can be denoted as:
\begin{equation}
\begin{gathered}
m=1, \\
\mathbf{\hat q}_s^{(1)}= \mathbf{q}_s,
\mathbf{\hat k}_t^{(1)}=\mathbf{k}_t,
\mathbf{W}_{t-s}^{(1)}=\mathbf W_{t-s}.
\end{gathered}
\end{equation}

\subsubsection{Simplification}
For the ease of the remaining discussion, we introduce the necessary notations and simplify Eq.~\ref{eq:rel}.
\begin{equation}
\begin{gathered}
\hat d_1=\sum_{l=1}^m d^{(l)}_1, \hat d_2=\sum_{l=1}^m d^{(l)}_2, \\
\hat {\mathbf q}_s = 
\left[\begin{matrix}
({\hat {\mathbf q}_s^{(1)}})\tran,\ldots,
({\hat{\mathbf q}_s^{(m)}})\tran
\end{matrix}\right]\tran\in \mathbb C^{\hat d_1},
\hat {\mathbf k}_t = 
\left[\begin{matrix}
({\hat {\mathbf k}_t^{(1)}})\tran,\ldots,
({\hat{\mathbf k}_t^{(m)}})\tran
\end{matrix}\right]\tran \in \mathbb C^{\hat d_2},\\
\hat {\mathbf  W}_{t-s}= \text{block-diag} \{ {\mathbf W}_{t-s}^{(1)}\ldots , 
{\mathbf  W}_{t-s}^{(m)}\}\in \mathbb C^{\hat d_1 \times \hat  d_2}.
\end{gathered}
\end{equation}
With these notations, we can rewrite Eq.~\ref{eq:rel} into the matrix form: $f_{\mathrm{rel}}(\mathbf q_s, \mathbf k_t)= {\hat{\mathbf  q}_s}\her \hat {\mathbf W}_{t-s} \hat{\mathbf  k}_t$.
Since every component of $\hat {\mathbf q}_s$ and $\hat {\mathbf k}_t$ are handled with no difference,
without losing generality,
we only discuss cases where $m=1$:  
\begin{equation}
f_{\mathrm{rel}}(\mathbf q_s,\mathbf  k_t)=\mathbf  q_s\her  \mathbf W_{t-s} \mathbf k_t.
\label{general1}
\end{equation}

\subsection{Linearized relative position encoding}
\label{linear}
Eq.~\ref{eq:rel} is a canonical form of relative positional encoding, meaning that its variants are applicable to vanilla transformers but not necessarily for linear ones.
To design relative encoding compatible with linear transformers, the attention computation has to respect the decomposibilty condition.
This additional condition leads to the linearized relative position encoding (\lrpe) family, defined as follows.

\newtheorem{de_lrpe}{Definition}[section]
\begin{de_lrpe}
A relative position encoding is called linearized relative position encoding (\lrpe), when the following holds:
\begin{equation}
\begin{aligned}
 \forall\mathbf  q_s,\mathbf  k_t \in \mathbb C^{d},  &f_{\mathrm{rel}}(\mathbf q_s,\mathbf  k_t)
= \mathbf q_s\her  \mathbf W_{t-s}\mathbf  k_t \\
= & (\mathbf M_s \mathbf q_s)\her(\mathbf M_t\mathbf k_t)
=\mathbf q_s\her\mathbf  M_s\her\mathbf  M_t\mathbf  k_t, 
\end{aligned}
\label{eq_lrpe}
\end{equation}
where $\mathbf q_s,\mathbf  k_t\in \mathbb C^d$, $\mathbf W_s,\mathbf M_s\in \mathbb C^{d\times d},\mathbf  W_{0}=\mathbf  I_d$.
\end{de_lrpe}
The assumption of $\mathbf W_0=\mathbf I_d$ implies that the interaction between tokens from the same position only depends on the content, which is reasonable enough that most encoding methods respect.
In its essence, Eq.~\ref{eq_lrpe} ensures the positional matrix is decomposable.
In this way, the query-key inner-product can be avoided in the attention computation.
{\label{lrpe_speed}Consequently, complexity of computing {\lrpe} is $O(n d^2)$, where $n$ is sequence length, $d$ is embedding dimension as Appendix \ref{proof1} shows in detail.}

We prove that Eq.~\ref{eq_lrpe} can be simplified based on the following proposition:
\newtheorem{lrpe_simp}[de_lrpe]{Proposition}
\label{lrpe_simp}
\begin{lrpe_simp}
Eq.~\ref{eq_lrpe} is equivalent to Eq.~\ref{main_eq} and $\mathbf W_t$ is Unitary matrix,
\begin{equation}
\label{main_eq} 
\mathbf W_{t-s} =\mathbf  W_s\her \mathbf W_t.
\end{equation}
\end{lrpe_simp}
\begin{proof}[Proof of Proposition ~\ref{lrpe_simp}]
\label{proof2}
According to the arbitrariness of $\mathbf q_s, \mathbf k_t$, Eq.~\ref{eq_lrpe} is equivalent to
\begin{equation}
\label{eq_eq}
  \mathbf W_{t-s}= \mathbf M_s\her \mathbf M_t .
\end{equation}
Take $s=t$ in Eq~\ref{eq_lrpe}, we get (since we assume that $\mathbf W_0= \mathbf I_d$):
\begin{equation}
 \mathbf M_s\her \mathbf M_s =\mathbf W_0= \mathbf I_d.
\end{equation}
Thus, $\mathbf M_s$ is a unitary matrix. On the other hand, note that for any unitary matrix $\mathbf P$, we always have
\begin{equation}
\begin{aligned}
  \mathbf W_{t-s}
  =& \mathbf M_s\her \mathbf M_t = \mathbf M_s\her \mathbf I_d \mathbf M_t \\
  =&\mathbf M_s\her \mathbf P\her \mathbf P\mathbf M_t=(\mathbf P\mathbf M_s)\her (\mathbf P \mathbf M_t).
\end{aligned}
\end{equation}
This means that left multiplying $\mathbf M_t$  by a unitary matrix $\mathbf P$ does not change  Eq.~\ref{eq_lrpe}. Since $\mathbf M_s$ and $\mathbf M_0\her$ are also unitary matrices, we can perform the following transformation:
\begin{equation}
\overline {\mathbf M}_s = \mathbf M_0\her \mathbf M_s.
\end{equation}
With $\overline{\mathbf M}_s$, Eq.~\ref{eq_eq} becomes
\begin{equation}
\label{eq_eq1}
\mathbf W_{t-s} = \overline{\mathbf M}_s\her \overline{\mathbf M}_t.
\end{equation}
Take $s=0$, we have
\begin{equation}
\mathbf W_t =  \overline{\mathbf M}_0\her \overline{\mathbf M}_t
= \mathbf M_0\her \mathbf M_0  \overline{\mathbf M}_t
=\mathbf I_d \overline{\mathbf M}_t=\overline {\mathbf M}_t.
\end{equation}
Thus Eq.~\ref{eq_eq1} becomes
\begin{equation}
\mathbf W_{t-s} = \mathbf W_s\her \mathbf W_t.
\end{equation}
Since $\overline {\mathbf M}_s$ is a unitary matrix, $\mathbf W_s$ is also a unitary matrix, \ig,
\begin{equation*}
\mathbf W_s\her \mathbf W_s = \mathbf I_d. \qedhere
\end{equation*}
\end{proof}
In the following section, we derive some particular solutions of Eq.~\ref{main_eq}.

\subsubsection{Particular solutions}
In this section, we discuss Eq.~\ref{main_eq} and give a family of solutions. It is worth noting that the solutions we provide are all in the form of $\mathbf W_s =\mathbf P\her \mathbf\Lambda^{(s)}\mathbf P$, where $\mathbf P, \mathbf \Lambda^{(s)}$ are unitary matrices. The complete derivation can be found in Appendix ~\ref{proof3}, ~\ref{proof4}, ~\ref{proof5}. 

\textbf{Unitary (Solution 1)}
The first case is discussed in the complex domain, which is not common in transformer models yet exhibiting an elegant solution.
\newtheorem{p1}[de_lrpe]{Proposition}

\begin{p1}
\label{p1}
The following form of {$\mathbf W_s \in \mathbb C^{d\times d}$} satisfies Eq.~\ref{main_eq}:
\begin{equation}
\begin{aligned}
\mathbf W_s &=\mathbf P\her \mathbf\Lambda^{(s)}\mathbf P,\\
\mathbf\Lambda^{(s)} &= \mathrm{diag} \{\exp(i s\alpha_1),\ldots, \exp(i s\alpha_d)\},\\
\end{aligned}
\end{equation}
where {$\mathbf P\in \mathbb C^{d\times d}$} is \textbf{unitary} matrix, $\alpha_k, k=1,\ldots,d$ are parameters.
\end{p1}

\textbf{Orthogonal (Solution 2)}
Now we consider the real domain, a more general case in transformers.
\newtheorem{p2}[de_lrpe]{Proposition}

\begin{p2}
\label{p2}
The following form of {$\mathbf W_s \in \mathbb R^{d\times d}$} satisfies Eq.~\ref{main_eq}:
\begin{equation}
\begin{gathered}
\mathbf W_s 
=\mathbf  P\tran \mathbf \Lambda^{(s)}\mathbf  P,
\mathbf \Lambda^{(s)} =\text{block-diag}\{\mathbf A^{(s)}, \mathbf B^{(s)} \},  \\
\mathbf A^{(s)}=\text{block-diag}\{\mathbf A_1^{(s)},\ldots, \mathbf A_n^{(s)} \} \in \mathbb R^{2p\times 2p}, 
\mathbf B^{(s)}= \mathbf I_q\in \mathbb R^{q\times q},\\
\mathbf A_k^{(s)}=\left[\begin{matrix}
\cos (s\alpha_k) & -\sin (s\alpha_k) \\ 
\sin (s\alpha_k) & \cos (s\alpha_k)
\end{matrix}\right],
\end{gathered}
\end{equation}
where {$\mathbf P\in \mathbb R^{d\times d}$} is \textbf{orthogonal} matrix, $\alpha_k, k=1,\ldots,d$ are parameters.
\end{p2}

\textbf{Permutation (Solution 3)}
The last case is inspired by PermuteFormer \citep{chen2021permuteformer}, which is associated with the permutation matrix:

\newtheorem{p3}[de_lrpe]{Proposition}

\begin{p3}
\label{p3}

The following form of {$\mathbf W_k \in \mathbb R^{d\times d}$} satisfies Eq.~\ref{main_eq}:
\begin{equation}
\begin{aligned}
\mathbf W_k &= \mathbf P\tran \mathbf \Lambda^{(k)}\mathbf  P,\\
\pi:\{1,2, \cdots, d\} &\rightarrow \{1,2, \cdots, d\}\text{ is permutation},\\
\mathbf \Lambda^{(k)}&= (\mathbf I)_{\pi^k},
\end{aligned}
\end{equation}
where {$\mathbf P\in \mathbb R^{d\times d}$} is the \textbf{orthogonal} matrix.
\end{p3}

\begin{table*}[t]
\caption{{Quantitative results of the Roberta model fine-tuned on the GLUE dataset. MNLI is reported by the match/mismatch splits. CoLA is reported by Matthews correlation coefficient. All the other tasks are measured by accuracy. The best result is highlighted with \textbf{bold} and the second with \underline{underlined}. $\downarrow$ means \textit{smaller is better}}.}
\vspace{2mm}
\label{blm_ft1}
    \centering
    \setlength{\tabcolsep}{3.92mm}
    \resizebox{\linewidth}{!} {
    \begin{tabular}{llllllllll}
    \hline
        Method & Loss$\downarrow$ &  MNLI &  QNLI &  QQP &  RTE &  SST-2 &  MRPC &  CoLA &  STS-B \\
        \hline
        Base & 5.35 & 76.10/76.39 & 85.47 & 88.44 & 53.43 & 89.33 & 70.59 & - & 48.08 \\ 
        Rope & 5.17 & 76.53/77.08 & 82.77 & 83.30 & 55.96 & \underline{90.48} & 69.36 & 38.84 & 49.28 \\ 
        SPE & 6.07 &  67.92/68.12 &73.70 & 87.27 & 53.43 & 84.75 & 70.34 & - & 17.88 \\ 
        PER & 5.32 & 77.30/77.37 & 84.09 & 89.03 & 55.96 & 90.25 & 71.08 & 29.04 & 68.10
        \\ \hline
        Type1 & 5.18 & \underline{79.18}/\underline{78.76} & \textbf{87.75} & \underline{89.55} & 55.96 & \underline{90.48} & \underline{73.04} & \underline{48.27} & \textbf{82.09} \\ 
        Type2 & \textbf{5.12} & \textbf{80.28}/\textbf{80.68} & 87.17 & \textbf{89.68} & \textbf{59.21} & \textbf{91.97} & \textbf{73.77} & \textbf{49.34} & \underline{79.28} \\ 
        Type3 & 5.28 & 77.34/77.55 & 86.22 & 88.99 & \underline{58.48} & \underline{90.48} & 71.08 & 36.67 & 74.66 \\ \hline
    \end{tabular}
}
\end{table*}
\subsection{The {\name} family}
\label{imple}

{\name} ($\mathbf W_s =\mathbf P\her \mathbf\Lambda^{(s)}\mathbf P$) contains two components, \ig, a fixed unitary matrix $\mathbf P$ and a unitary matrix family $\mathbf \Lambda^{(s)}$ as mentioned in proposition \ref{p1}, \ref{p2}, and \ref{p3}. 
The $\mathbf P$ can be seen as a rotation matrix that rotates the token feature to a particular coordinate system and the $\mathbf \Lambda^{(s)}$ derives the positional information from the rotated feature. 

To meet all the requirements in proposition \ref{p1}, \ref{p2}, and \ref{p3}, $\mathbf P$ needs to be an orthogonal matrix. We empirically find that when $\mathbf P$ is a householder matrix~\citep{golub2013matrix}, the overall performance is better than other options such as permutation matrix and Identity matrix. We provide a detailed ablation in Table~\ref{ablatP}. For ease of expression, we use \emph{Type 1} for the unitary solution, \emph{Type 2} for the orthogonal solution, and \emph{Type 3} for the permutation solution.
Details can be found in Appendix \ref{imp_detail}.

\section{Experiments}

In this section, we validate the effectiveness of the proposed {\name} on natural language processing tasks and computer vision tasks that resort to different Transformer architectures. Specifically, we first study the autoregressive language model~\citep{radford2018improving}. This is followed by the bidirectional language model, which adopts the Roberta architecture~\citep{liu2020roberta} and is pretrained and then fine-tuned on several downstream tasks from the GLUE benchmark \citep{wang2018glue}. We also extend our evaluation on image classification task to verify the generalization ability of \name. 

\subsection{Experimental settings}
\paragraph{Dataset}
We use Wikitext-103~\cite{1609.07843}, Books~\cite{Zhu_2015_ICCV}, and WikiBook~\cite{2202.08005} datasets for NLP task evaluation and ImageNet-1k~\cite{deng2009imagenet} for image classification evaluation. Wikitext-103 is a small dataset containing a preprocessed version of the Wikipedia dataset. Books consists of a large number of novels, making it suitable for long sequence modeling evaluation. WikiBook is a large corpus (22 GB) of Wikipedia articles and books collected by~\cite{2202.08005}.
ImageNet-1k is the most popular large image classification dataset. It contains 1000 object classes and over 1 million training images and is often used to verify the performance of models in image modeling.

\vspace{-1.5mm}
\paragraph{Configurations} Our experiments are implemented in the \textit{Fairseq} framework \citep{ott2019fairseq} and trained with V100 GPUs. All the methods share the same configurations such as learning rate, batch size, and optimizer. The detailed configurations are listed in Appendix ~\ref{config}.

\begin{figure*}[t]
  \begin{center}
  {\includegraphics[width=0.48\linewidth, trim=0 0 0 0]{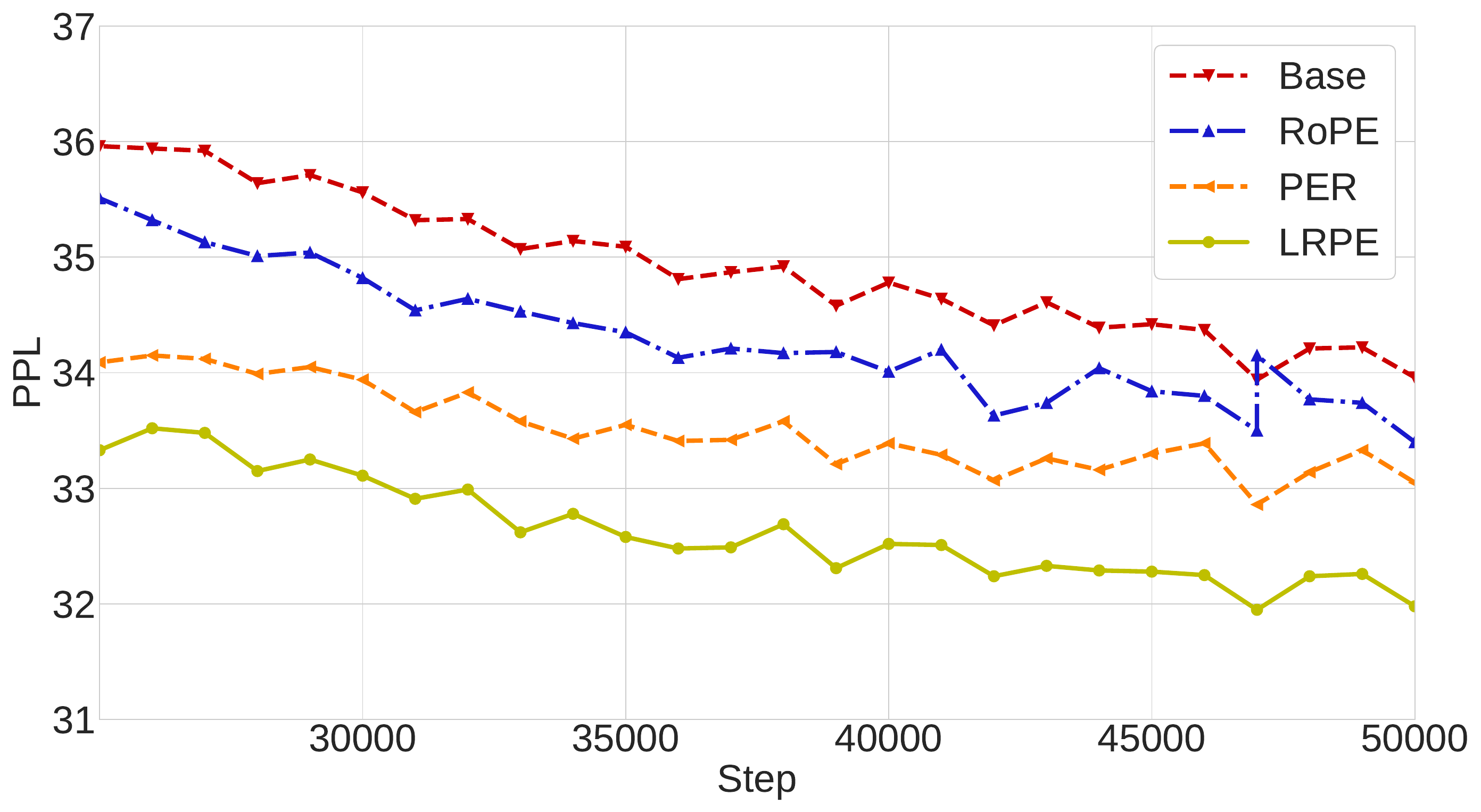}
  \includegraphics[width=0.48\linewidth, trim=0 0 0 0]{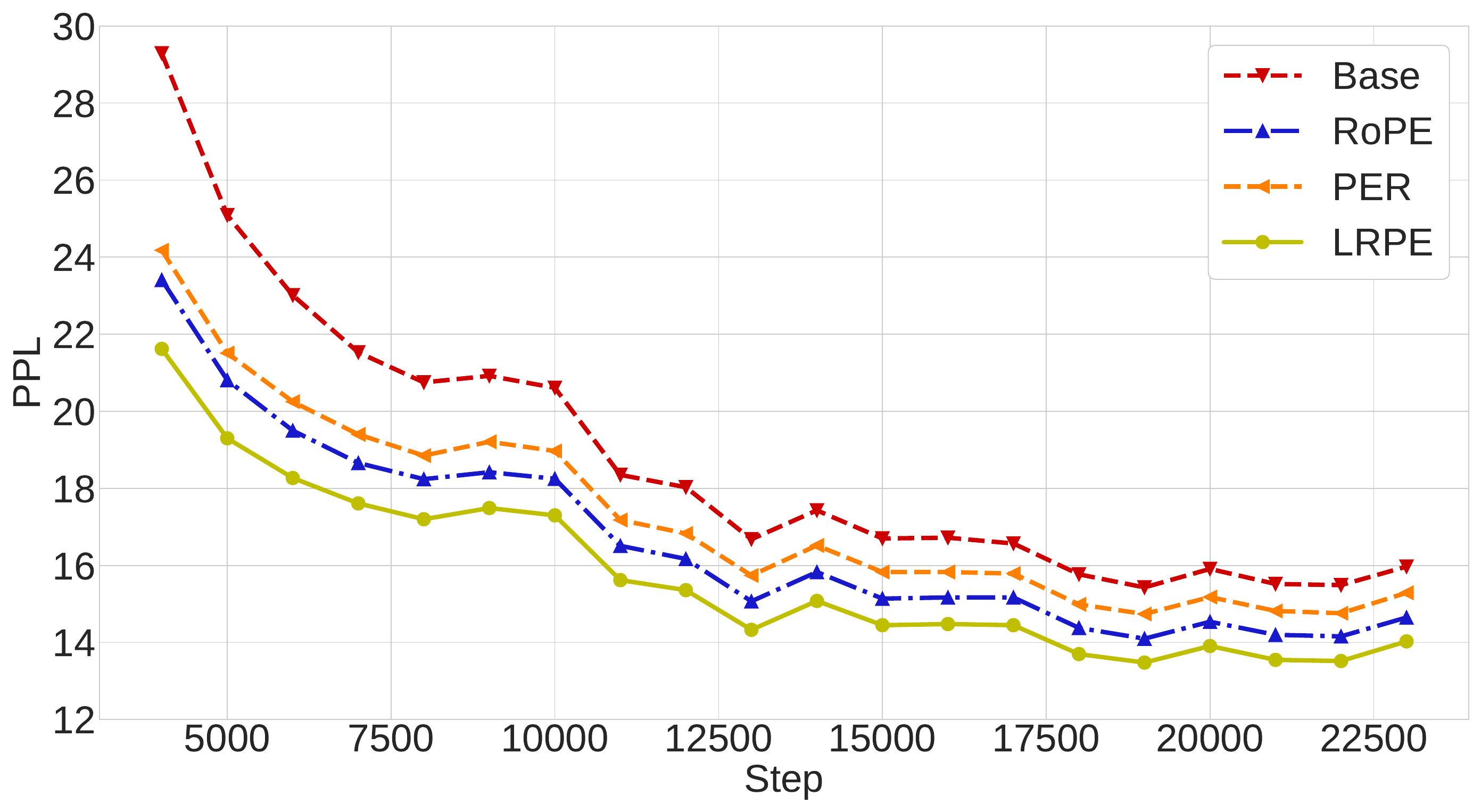}} 
   \end{center}
   \label{fig: bi ppl}
   \vspace{-2mm}
\caption{\small The validation result of Autoregressive language model on the Wikitext-103 dataset(left) and Roberta on the Wikibook dataset(right). In both cases, the best result of the proposed {\name} has a better PPL and faster convergence speed than competing methods.
}
\end{figure*}

\vspace{-1.5mm}
\paragraph{Competing methods} Our baseline (marked as Base) is a Linear Transformer with $1+\mathrm{elu}(\cdot)$ \citep{katharopoulos2020transformers} as the kernel function with sinusoidal positional encoding~\citep{vaswani2017attention}.
For comparison, we also choose several state-of-the-art methods, \ig, RoPE \citep{su2021roformer}, SPE \citep{liutkus2021relative}, PermuteFormer (abbreviated as ``PER'') \citep{chen2021permuteformer}.

\subsection{Results}
\begin{wraptable}[13]{r}{.5\linewidth}
\small
\vspace{-7.8mm}
\caption{Quantitative results of the autoregressive language model on the WikiText-103 and Books dataset. The best result is highlighted with \textbf{bold} and the second best with \underline{underlined}. $\downarrow$ means \textit{smaller is better}.}
 \centering
 \setlength{\tabcolsep}{4.0mm}
 \label {lm_res}
    \begin{tabular}{lllll}
    \hline
        {\multirow{2}{*}{Method}} & \multicolumn{2}{c}{\textbf{Wikitext-103}} & \multicolumn{2}{c}{\textbf{Books}}  \\ \cline{2-5}
        ~ & Val$\downarrow$ & Test$\downarrow$ & Val$\downarrow$ & Test$\downarrow$\\ 
        \hline
        Base & 33.94 & 33.74 & 9.12 & 8.78 \\ 
        Rope & 33.40 & 33.13 & 8.98 & 8.65 \\ 
        SPE & 43.50 & 41.91 & 11.91 & 10.85 \\ 
        PER & 32.86 & 32.53 & 8.53 & 8.20 \\ \hline
        Type1 & \textbf{31.90} & \textbf{31.60} & 
        \underline{8.52} & \underline{8.19} \\ 
        Type2 & \underline{31.95} & \underline{31.71} & \textbf{8.46} & \textbf{8.14} \\ 
        Type3 & 33.90 & 33.95 & 8.69 & 8.43 \\ \hline
    \end{tabular}
\end{wraptable}
\textbf{Autoregressive language model.} The autoregressive language model has 6 decoder layers and is trained on the WikiText-103 dataset \citep{merity2017pointer}. In order to test the performance of the method on long sequence modeling, we tested the performance of the model on the Books~\citep{Zhu_2015_ICCV} dataset. We use the Perplexity (PPL) as the evaluation metric and report the results in Table~\ref{lm_res}. 
We observe that all variants of {\name} present a performance gain over the baseline. Notably, Type 1 and Type 2 models achieve the best performance on Wikitext-103 and Books, respectively, demonstrating superior capability in language modeling. 

\begin{wraptable}[12]{r}{.5\linewidth}
    \centering
    \small
    \vspace{-7.8mm}
\caption {Quantitative results of image classification on the ImageNet-1k dataset. The best result is highlighted with \textbf{bold} and the second with \underline{underlined}. $\uparrow$ means \textit{larger is better}}
\vspace{2mm}
\label{cvtable}
    \centering
    \setlength{\tabcolsep}{8mm}
    \begin{tabular}{lll}
    \hline
        Method & Acc$\uparrow$ & Params \\ \hline
        Base & 77.88 & 22.04 \\ 
        RoPE & 78.52 & 22.04 \\ 
        PER & 77.99 & 22.04 \\ \hline
        Type1 & \underline{78.58} & 22.05 \\ 
        Type2 & \textbf{78.78} & 22.05 \\
        Type3 & 77.74 & 22.05 \\ \hline
    \end{tabular}
\end{wraptable}
\textbf{Bidirectional language model.} The bidirectional model follows an encoder-only structure, \ig, Roberta \citep{liu2020roberta}, with 12 layers. 
In order to verify the performance of the model on a large data set, we adopt the Wikibook dataset used by~\cite{2202.08005}  for pre-training and used their configurations to update 23k times. The results are in Table~\ref{blm_ft1} and Figure~\ref{fig: bi ppl}. In the pre-training phase, \name~outperforms all competitors.
Next, we fine-tune the model for the GLUE task. 
As shown in Table~\ref{blm_ft1}, our method outperforms competing methods on all tasks with a clear margin.

\textbf{Image classification model. }
To verify the robustness and effectiveness of \name~ under different modal tasks, we test our method on the computer vision domain. Specifically, we conduct experiments on Imagenet-1k~\cite{deng2009imagenet} dataset using the Deit-small architecture~\cite{pmlr-v139-touvron21a} on the image classification task. In particular, we replace the Attention with Linear Attention~\cite{katharopoulos2020transformers} and then adopt various relative positional encoding. As shown in Table~\ref{cvtable}, LRPE beats all the competing methods.

\textbf{Long-Range Arena. } In order to validate the effectiveness of LRPE on long-sequence modeling tasks, we conducted experiments on Long-Range Arena benchmark~\citep{tay2020long}. As shown in Table~\ref{lra}, {\name} has positive effects on almost all tasks.

\begin{table}[!ht]
\caption {Quantitative results of classification tasks on the Long-Range Arena benchmark. The best result is highlighted with \textbf{bold} and the second with \underline{underlined}. $\uparrow$ means \textit{larger is better}.}
    \centering
    \setlength{\tabcolsep}{5.5mm}
    \label{lra}
    \begin{tabular}{lllllll}
    \hline
        Model & Text &  ListOps  & Retrieval  & Pathfinder  & Image & AVG \\ \hline
        Base & 64.81  & 39.17  & \textbf{84.71}  & 71.61  & 41.21  & 60.30  \\ 
        RoPE & \underline{66.53}  & \textbf{40.57}  & 78.75  & 72.68  & \underline{62.50}  & 64.59  \\ 
        Per & 65.50  & 39.17  & 83.40  & 69.55  & 42.03  & 60.86  \\ \hline
        Type1 & \textbf{66.80}  & \underline{39.84}  & 82.13  & \textbf{73.79}  & \textbf{63.43}  & \textbf{65.47}  \\
        Type2 & 66.52  & 39.79  & 81.55  & \underline{73.46}  & 62.35  & \underline{65.03}  \\ 
        Type3 & 64.94  & 39.56  & \underline{83.86}  & 70.78  & 45.15  & 61.54 \\ \hline
    \end{tabular}
\end{table}

\subsection{Discussion}
\paragraph{An explanation of {\name}. }

\begin{wraptable}[8]{r}{.54\linewidth}
  \small
  \vspace{-8mm}
\caption{Ablation results with different rotation matrix $\mathbf{P}$ for language modeling on the WikiText-103 dataset.}
    \vspace{2mm}
    \label{rotate}
    \centering
    \setlength{\tabcolsep}{2mm}
    \begin{tabular}{lcccc}
    \hline
       \diagbox{$\mathbf P$ }{$\mathbf \Lambda^{(s)}$} & Unitary & Orthogonal & Permutation & Avg. \\ \hline
        Householder & 31.90 & 31.95 & 33.90 & 32.58 \\ 
        Identity & 32.04 & 31.86 & 34.53 & 32.80 \\ 
        Permutation & 32.09 & 31.59 & 34.16 & 32.61 \\ \hline
    \end{tabular}
    \label{ablatP}
\end{wraptable}

According to the discussion in Section.~\ref{imple}, The proposed {\name} rotates the token feature through $\mathbf P$, and encodes the positional information through $\mathbf \Lambda^{(s)}$ . In Table~\ref{rotate}, we ablate the effectiveness of the $\mathbf P$ matrix on the autoregressive language modeling task.
Our approach with the Householder matrix achieves better results than the one equipped with other metrics. It indicates that we can get better performance by carefully selecting the projection of the positional encoding. 

\begin{wraptable}[11]{r}{.54\linewidth}
    \centering
  \small
  \vspace{-7.8mm}
\caption{Training speed of different methods on the bidirectional language model. The value standards for the speed relative to the base method. $\uparrow$ means \textit{larger is faster}. }
\vspace{1mm}
\label{speed}
    \centering
    \setlength{\tabcolsep}{14mm}
    \begin{tabular}{ll}
    \hline
        Method & Relative speed$\uparrow$
    \\ \hline
        Base  & \makecell{1.00}  \\
        Rope & \makecell{0.86} \\ 
        SPE & \makecell{0.61} \\
        PER & \makecell{0.94} \\
        \hline
         Type1 & \makecell{0.82} \\
         Type2 & \makecell{0.82} \\
          Type3 & \makecell{0.89} \\
        \hline
    \end{tabular}
\end{wraptable}

\paragraph{Complexity and efficiency. }The implementation of the proposed {\name} does not affect the computational complexity of the linear transformer, \ig, preserving the linear complexity as $O(n)$. We also measure the training speed of the bidirectional language model on the same local machine and observe that the speed after using {\name} is only a bit slower than the baseline on average. The detailed comparison of the efficiency can be found in Table \ref{speed}. In general, \name~does not incur a significant computational burden to the transformer, and can fulfill the practical needs by maintaining comparable efficiency. 

\section{Conclusion}
In this paper, we standardize the form of relative positional encoding for linear attention. The unitary transformation is employed as a special solution to the linearized relative positional encoding, and the solutions as per various constraints constitute the unitary relative positional encoding ({\name}) family. We validate the effectiveness of {\name} through extensive experiments on both natural language processing and computer vision tasks with different transformer architectures. It outperforms competing methods in all tasks. In addition, it highlights a broad paradigm for formulating linear transformer-applicable positional encoding techniques that are more generically relative.

\bibliography{main}
\bibliographystyle{tmlr}

\appendix
\newpage
\onecolumn
% \section{Appendix}
\begin{center}
\textbf{\large Appendix}
\end{center}

%%%%%%%%%% Merge with supplemental materials %%%%%%%%%%
% \pagebreak
% \widetext
% \widetext
% \clearpage
% \newpage
% \resetlinenumber
% \begin{center}
% \textbf{\large Supplemental Materials: Title for main text}
% \end{center}
%%%%%%%%%% Merge with supplemental materials %%%%%%%%%%
%%%%%%%%%% Prefix a "S" to all equations, figures, tables and reset the counter %%%%%%%%%%
% \setcounter{equation}{0}
% \setcounter{figure}{0}
% \setcounter{table}{0}
% \setcounter{page}{1}
% \makeatletter
% \renewcommand{\theequation}{S\arabic{equation}}
% \renewcommand{\thefigure}{S\arabic{figure}}
% \renewcommand{\bibnumfmt}[1]{[S#1]}
% \renewcommand{\citenumfont}[1]{S#1}

%%%%%%%%%% Prefix a "S" to all equations, figures, tables and reset the counter %%%%%%%%%%

\section{Mathematical Notations}
\label{math_notation}

\begin{table}[!ht]
\centering
\small
\setlength{\tabcolsep}{15.95mm}{
\begin{tabular}{cc}
\hline\hline
\textbf{Notation} & \textbf{Meaning} \\
\hline\hline
$\mathbf{X}$ & Hidden state. \\
$\mathbf{Q},\mathbf{K},\mathbf{V}$ & Query, key, value. \\
$\mathbf{W}_Q,\mathbf{W}_K,\mathbf{W}_{V}$ & Weight matrices for $\mathbf{Q}$,$\mathbf{K}$,$\mathbf{V}$. \\
$\mathbf{O}$ & Attention output. \\
$\mathbf{m}\tran_s$ & $s$-th row of matrix $M$ (real domain). \\
$\mathbf{m}\her_s$ & $s$-th row of matrix $M$ (complex domain). \\
$\phi$ & Kernel function for linear attention. \\
$\mathbf 1_d $ & All-ones vector with dimention $d$. \\
$\mathbf I_d $  & Identity matrix with dimention $d$. \\
block-diag &  Combining matrices into larger \\
    &block diagonal matrices as in Eq.~\ref{block_diag}\\
\hline\hline
\end{tabular}}
\caption{Mathematical notations used in the paper.}
\end{table}

\begin{equation}
\label{block_diag}
\text{block-diag} \{\mathbf{W}_1,\mathbf{W}_2,\ldots,\mathbf{W}_n\}=
 \left[
 \begin{matrix}
   \mathbf{W}_1 &     &        \\
       & \mathbf{W}_2 &   \\
       &     &  \ddots  \\
       &     &   & \mathbf W_n
  \end{matrix}
  \right].
\end{equation}
\section{Computation of Vanilla/Linear Attention}
\label{compute}
% In this section, we give some basic notation and then discuss the computation of vanilla and linear attention.
\subsection{Basic Notations}
Both vanilla and linear attention blocks involve three matrices, \ig, $\mathbf Q$ (\textbf{Query}), $\mathbf K$ (\textbf{Key}) 
and $\mathbf V$ (\textbf{Value}). All of them are linear projections of input $\mathbf X\in \mathbb C^{n\times d}$, \ig, 
\begin{equation}
\begin{aligned}
\mathbf{X}
&=\left[
 \begin{matrix}
\mathbf{x}_1\tran \\
\vdots\\
\mathbf{x}_n\tran
  \end{matrix}
  \right] \in \mathbb R^{n\times d},\\
\mathbf{Q}
&=\left[
 \begin{matrix}
\mathbf{q}_1\tran\\
\vdots\\
\mathbf{q}_n\tran
  \end{matrix}
  \right] =\mathbf{XW}_Q=\left[
 \begin{matrix}
\mathbf{x}_1\tran\mathbf{W}_Q\\
\vdots\\
\mathbf{x}_n\tran\mathbf{W}_Q
  \end{matrix}
  \right] \in \mathbb R^{n\times d}, \\
 \mathbf{K}&= \left[
 \begin{matrix}
 \mathbf{k}_1\tran\\
\vdots\\
 \mathbf{k}_n\tran
  \end{matrix}
  \right] 
  =\mathbf{XW}_K=\left[
 \begin{matrix}
\mathbf{x}_1\tran\mathbf{W}_K\\
\vdots\\
\mathbf{x}_n\tran\mathbf{W}_K
  \end{matrix}
  \right] \in \mathbb R^{n\times d}, \\
\mathbf{V}&=\left[
 \begin{matrix}
\mathbf{v}_1\tran\\
\vdots\\
\mathbf{v}_n\tran
  \end{matrix}
  \right] 
  =\mathbf{XW}_V=\left[
 \begin{matrix}
\mathbf{x}_1\tran\mathbf{W}_V\\
\vdots\\
\mathbf{x}_n\tran\mathbf{W}_V
  \end{matrix}
  \right] \in \mathbb R^{n\times d}, 
\end{aligned}
\end{equation}
where $\mathbf{W}_Q,\mathbf{W}_K,\mathbf{W}_V \in \mathbb{R}^{d\times d}$.

The vector form is organized as
\begin{equation}
\begin{gathered}
  \mathbf{q}_s = \mathbf{W}_Q\tran  \mathbf{x}_s,
 \mathbf{k}_s =\mathbf{W}_K\tran \mathbf{x}_s,
\mathbf{v}_s =\mathbf{W}_V\tran \mathbf{x}_s.
\end{gathered}
\end{equation}
The attention output is 
\begin{equation}
\mathbf{O}=\left[
 \begin{matrix}
\mathbf{o}_1\tran\\
\vdots\\
\mathbf{o}_n\tran
  \end{matrix}
  \right] \in \mathbb  R^{n\times d}.
\end{equation}

\subsection{Vanilla Attention}
\label{vanilla_attn}

In vanilla attention, the output is computed using the Softmax weighted sum, \ig, 
\begin{equation}
\begin{aligned}
\mathbf{o}_s
&=\textrm{Attention}(\mathbf{q}_s, \mathbf{K}, \mathbf{V})\\
&=\sum_{t=1}^n \mathbf{a}_{st}  \mathbf{v}_t\\
&= \sum_{t=1}^n \frac{\exp\left(\mathbf{q}_s\tran\mathbf{k}_{t}/\sqrt{d} \right) \mathbf v_t}
{\sum_{r=1}^n\exp\left(\mathbf{q}_s\tran\mathbf{k}_{r}/\sqrt{d} \right)},\\
\mathbf O&= \textrm{Softmax}(\mathbf{Q K}\tran / \sqrt{d}) \mathbf{V}.
\end{aligned}
\end{equation}

\subsection{Linear Attention}
\label{linear_attn}
The linear attention is formulated as follows,

\begin{equation}
\begin{aligned}
\mathbf{o}_s
&=\textrm{LinearAttention}(\mathbf{q}_s, \mathbf{K}, \mathbf{V})\\
&=\sum_{t=1}^n \mathbf{a}_{st}  \mathbf{v}_t\\
&=\sum_{t=1}^n \frac{\phi(\mathbf{q}_s)\tran \phi(\mathbf{k}_{t})}
{\sum_{t=1}^n\phi(\mathbf{q}_s)\tran \phi(\mathbf{k}_{t})}  \mathbf{v}_t\\
&= \frac{\sum_{t=1}^n\phi(\mathbf{q}_s)\tran \phi(\mathbf{k}_{t}) \mathbf{v}_t}
{\sum_{t=1}^n\phi(\mathbf{q}_s)\tran \phi(\mathbf{k}_{t})} \\
&= \phi(\mathbf{q}_s)\tran\frac{\sum_{t=1}^n \phi(\mathbf{k}_{t}) \mathbf{v}_t}
{\phi(\mathbf{q}_s)\tran\sum_{t=1}^n \phi(\mathbf{k}_{t})}, \\
\mathbf{O}&=\mathbf \Delta^{-1}  \phi(\mathbf{Q}) \phi(\mathbf{K})\tran\mathbf{V} \\
&=\mathbf \Delta^{-1} \phi(\mathbf{Q}) [\phi(\mathbf{K})\tran \mathbf{V} ],\\
\mathbf \Delta& = \mathrm{diag}(\phi(\mathbf{Q})  [\phi(\mathbf{K})\tran {\mathbf 1}_n]).
\end{aligned}
\end{equation}

\section{Proof of Theorem}
\label{proof_theorem}
\subsection{More Examples}
In the following, we provide two additional examples of relative positional encoding with the canonical form.
\label{srpe}

% \textbf{RPR} \citep{shaw-etal-2018-self}:
% \begin{equation}
% \begin{aligned}
% f_{\mathrm{rel}}(\mathbf{q}_s, \mathbf{k}_t)
% &=\mathbf{q}_s\her \mathbf{k}_t+\mathbf{q}_s\her \mathbf{c}_{t-s},  \\
%  \mathbf{c}_{t-s} &= \mathbf{w}_{\mathrm{clip}(t-s, k)}, \\
%  \mathrm{clip}(x, k)&=\max (-k, \min (k, x)), \\
%  \mathbf{w_{s}}&\in \mathbb C^ d ,-k\le s \le k .
% \end{aligned}
% \end{equation}
% The canonical form is
% \begin{equation}
% \begin{gathered}
% m=2, \\
% \mathbf{\hat q}_s^{(1)}= \mathbf{ q}_s,\mathbf{\hat k}_t^{(1)}=\mathbf{ k}_t,\mathbf{W}_{t-s}^{(1)}=\mathbf{I}_d,\\
% \mathbf{\hat q}_s^{(2)}=\mathbf{q}_s,\mathbf{\hat k}_t^{(2)}= \mathbf{I}_d,\mathbf{W}_{t-s}^{(2)}=\frac 1 d  
% \underbrace{\left[
%  \begin{matrix}
%   \mathbf{c}_{t-s} & \ldots & \mathbf{c}_{t-s}
%   \end{matrix}
%   \right]}_{d\text{ columns}}.
% \end{gathered}
% \end{equation}

\textbf{DeBERTa} \citep{huang-etal-2020-improve}:
\begin{equation}
\begin{aligned}
f_{\mathrm{rel}}(\mathbf{q}_s, \mathbf{k}_t)
&=\mathbf{q}_s\her \mathbf{k}_t+\mathbf{q}_s\her \mathbf{\bar k}_{{ g(s-t)}} +\mathbf{\bar q}_{{ g(t-s)}}\her \mathbf{k}_t, \\
 g(x)&=\begin{cases}
0 & x \le-c \\
2 c-1 & x \ge c \\
x+c & \text {others}.
\end{cases}
\end{aligned}
\end{equation}
The canonical form is
\begin{equation}
\begin{gathered}
m=3, \\
\mathbf{\hat q}_s^{(1)}= \mathbf{q}_s,\mathbf{\hat k}_t^{(1)}=\mathbf{k}_t,\mathbf{W}_{t-s}^{(1)}=\mathbf{I}_d,\\
\mathbf{\hat q}_s^{(2)}=\mathbf{q}_s,\mathbf{\hat k}_t^{(2)}= \mathbf{I}_d,\mathbf{W}_{t-s}^{(2)}=\frac 1 d  
\underbrace{\left[
 \begin{matrix}
  \mathbf{\bar  k}_{{ g(s-t)}} & \ldots & \mathbf{\bar k}_{{ g(s-t)}}
  \end{matrix}
  \right]}_{d\text{ columns}}, \\
\mathbf{\hat q}_s^{(3)}=\mathbf{I}_d
 ,\mathbf{\hat k}_t^{(3)}= \mathbf{k}_t,\mathbf{W}_{t-s}^{(3)}=\frac 1 d  
\underbrace{\left[
 \begin{matrix}
 \mathbf{\bar q}_{{g(t-s)}} & \ldots & \mathbf{\bar q}_{{ g(t-s)}}
  \end{matrix}
  \right]}_{d\text{ columns}}.
\end{gathered}
\end{equation}

\textbf{RPR}~\citep{shaw-etal-2018-self}:
\begin{equation}
\begin{aligned}
f_{\mathrm{rel}}(\mathbf{q}_s, \mathbf{k}_t)
&=\mathbf{q}_s\her \mathbf{k}_t+\mathbf{q}_s\her \mathbf{c}_{t-s},  \\
 \mathbf{c}_{t-s} &= \mathbf{w}_{\mathrm{clip}(t-s, k)}, \\
 \mathrm{clip}(x, k)&=\max (-k, \min (k, x)), \\
 \mathbf{w_{s}}&\in \mathbb C^ d ,-k\le s \le k .
\end{aligned}
\end{equation}
The canonical form is
\begin{equation}
\begin{gathered}
m=2, \\
\mathbf{\hat q}_s^{(1)}= \mathbf{ q}_s,\mathbf{\hat k}_t^{(1)}=\mathbf{ k}_t,\mathbf{W}_{t-s}^{(1)}=\mathbf{I}_d,\\
\mathbf{\hat q}_s^{(2)}=\mathbf{q}_s,\mathbf{\hat k}_t^{(2)}= \mathbf{I}_d,\mathbf{W}_{t-s}^{(2)}=\frac 1 d  
\underbrace{\left[
 \begin{matrix}
  \mathbf{c}_{t-s} & \ldots & \mathbf{c}_{t-s}
  \end{matrix}
  \right]}_{d\text{ columns}}.
\end{gathered}
\end{equation}

\textbf{cosFormer}~\citep{zhen2022cosformer}:
\begin{equation}
f_{\mathrm{rel}}(\mathbf{q}_s, \mathbf{k}_t)= \mathbf{q}_s\her \mathbf{k}_t\cos(\alpha(t-s)), 
\end{equation}
which indicates that the relative positional encoding is effectively a coefficient term in the attention matrix, as such, it can be derived via a positional matrix primitive with the coefficients.
\begin{equation}
\begin{gathered}
m=1, \\
\mathbf{\hat q}_s^{(1)}= \mathbf{q}_s,
\mathbf{\hat k}_t^{(1)}=\mathbf{k}_t,
\mathbf{W}_{t-s}^{(1)}=\cos(\alpha(t-s)) \mathbf{I}_d.
\end{gathered}
\end{equation}

% \textbf{cosFormer}~\citep{zhen2022cosformer}:
% \begin{equation}
% f_{\mathrm{rel}}(\mathbf{q}_s, \mathbf{k}_t)= \mathbf{q}_s\her \mathbf{k}_t\cos(\alpha(t-s)), 
% \end{equation}
% which indicates that the relative positional encoding is effectively a coefficient term in the attention matrix, as such, it can be derived via a positional matrix primitive with the coefficients.
% \begin{equation}
% \begin{gathered}
% m=1, \\
% \mathbf{\hat q}_s^{(1)}= \mathbf{q}_s,
% \mathbf{\hat k}_t^{(1)}=\mathbf{k}_t,
% \mathbf{W}_{t-s}^{(1)}=\cos(\alpha(t-s)) \mathbf{I}_d.
% \end{gathered}
% \end{equation}

% \subsection{Mathematical notation explained}
% \label{math_explained}
% Block-diag means combining matrices into larger block diagonal matrices:
% \begin{equation}
% \label{block_diag}
% \text{block-diag} \{\mathbf{W}_1,\mathbf{W}_2,\ldots,\mathbf{W}_n\}=
%  \left[
%  \begin{matrix}
%   \mathbf{W}_1 &     &        \\
%       & \mathbf{W}_2 &   \\
%       &     &  \ddots  \\
%       &     &   & \mathbf W_n
%   \end{matrix}
%   \right] 
% \end{equation}

\subsection{Speed analysis}
% \begin{proof}[Proof of Theorem~\ref{th_lrpe}]
\begin{proof}[Proof of Lrpe speed]
\label{proof1}
For this, we only need to prove that the time complexity is linear with respect to $n$. To this end, we first give basic notations as follows,
\begin{equation}
\begin{aligned}
 \mathbf Q= \left[
 \begin{matrix}
\mathbf{q}_1\her\\
\vdots\\
\mathbf{q}_n\her
  \end{matrix}
  \right] \in \mathbb  C^{n\times d},
 \mathbf K= \left[
 \begin{matrix}
 \mathbf k_1\her\\
\vdots\\
 \mathbf k_n\her
  \end{matrix}
  \right] \in \mathbb  C^{n\times d},
 \mathbf V=\left[
 \begin{matrix}
 \mathbf v_1\her\\
\vdots\\
 \mathbf v_n\her
  \end{matrix}
  \right] \in \mathbb  C^{n\times d}, \\
  \tilde{\mathbf{Q}}= \left[
 \begin{matrix}
(\mathbf M_1 \mathbf q_1)\her\\
\vdots\\
(\mathbf M_n \mathbf q_n)\her
  \end{matrix}
  \right] \in \mathbb  C^{n\times d},
\tilde {\mathbf K}= \left[
 \begin{matrix}
(\mathbf M_1 \mathbf k_1)\her\\
\vdots\\
(\mathbf M_n \mathbf k_n)\her
  \end{matrix}
  \right] \in \mathbb  C^{n\times d}.
\end{aligned}
\end{equation}

The time complexity of transforming $\mathbf Q, \mathbf K$ to $\tilde {\mathbf Q}, \tilde {\mathbf K}$ is $O(nd^2)$. The next step is to calculate the output, \ig, 
\begin{equation}
\begin{aligned}
\mathbf O&= \mathbf Q(\mathbf K\her \mathbf V) \in \mathbb C^{n\times d },\\
\mathbf O&=\mathbf \Delta^{-1}  \tilde {\mathbf Q} \tilde {\mathbf K}\her \mathbf V \\
&=\mathbf \Delta^{-1} \tilde {\mathbf Q} [\tilde {\mathbf K}\her \mathbf V ],\\
\mathbf \Delta& = \mathrm{diag}(\tilde {\mathbf Q} ) [\tilde {\mathbf K}\her {\mathbf 1}_n].
\end{aligned}
\label{sss}
\end{equation}
Clearly, Eq.~\ref{sss} is a standard formulation for the linear attention with the time complexity as $O(nd^2)$. Combing it with the first step, we have the total time complexity as $O(nd^2)$, which is unchanged.
\end{proof}

% \begin{proof}
% \label{proof2}
% According to the arbitrariness of $q_s, k_t$, equation \eqref{eq_slrpe} is equivalent to
% \begin{equation}
% \label{main_eq}
% W_{s}\her W_{t}= W_{t-s}
% \end{equation}
% Take $s=t$, we get
% \begin{equation}
% W_s\herW_s = W_0 \label{eq_uni}
% \end{equation}
% Take $s=0$, we get
% \begin{equation}
% W_0\herW_0= W_0
% \end{equation}
% Since $W_0$ is \textbf{invertible}, we get
% \begin{equation}
% W_0\her = I\Longleftrightarrow W_0 = I\her = I
% \end{equation}
% Take the result into the equation \eqref{eq_uni}, finnally we get
% \begin{equation}
% W_s\her W_s = I
% \end{equation}
% \end{proof}
% Prior to solving the problem, some theorems are clarified in the following \citep{Advanced.Algebra}:

%%%%%%%%%%%%%%%%
\subsection{Linearized Relative Positional Encoding}
\label{der_urpe}
Before the proof, we first give the following theorems \citep{Advanced.Algebra}:
\newtheorem{th_unitary_standard}[de_lrpe]{Theorem}
\begin{th_unitary_standard}
\label{th1}
If matrix $\mathbf W\in \mathbb C^{d\times d}$ is a unitary matrix, there exists another \textbf{unitary} matrix $\mathbf P\in \mathbb C^{d\times d}$, such that
\begin{equation}
\begin{aligned}
\mathbf W &=\mathbf  P\her \Lambda \mathbf P,\\
 \mathbf \Lambda &=\mathrm{diag} \{\exp(i\theta_1),\ldots, \exp(i\theta_d)\},\\
  i^2&= -1.
\end{aligned}
\end{equation}
\end{th_unitary_standard}

\newtheorem{th_or_standard}[de_lrpe]{Theorem}

\begin{th_or_standard}
\label{th2}
If matrix $\mathbf W\in \mathbb R^{d\times d}$ is an orthogonal matrix, there exists another \textbf{orthogonal} matrix $\mathbf P\in \mathbb R^{d\times d}$, such that
\begin{equation}
\begin{aligned}
\mathbf W &= \mathbf P\tran \mathbf \Lambda \mathbf P,\\
 \mathbf\Lambda &=\mathrm{diag} \{\mathbf\Lambda_1,\ldots,\mathbf\Lambda_r; 1,\ldots, 1; -1,\ldots,-1\},\\
\mathbf\Lambda_k&=\left[\begin{matrix}
\cos \theta_{k} & -\sin \theta_{k} \\
\sin \theta_{k} & \cos \theta_{k}
\end{matrix}\right], k=1,\ldots r.
\end{aligned}
\end{equation}
\end{th_or_standard}

\subsection{Unitary (Solution 1)}
\begin{proof}[Proof of Proposition~\ref{p1}]
\label{proof3}
According to Theorem {\ref{th1}}, we can assume that $\mathbf W_s$ has the following form ({$\mathbf P\in \mathbb C^{d\times d}$} is a \textbf{unitary} matrix),
\begin{equation}
\begin{aligned}
\mathbf W_s &= \mathbf P\her \mathbf \Lambda^{(s)} \mathbf P, \\
  \mathbf\Lambda^{(s)}&=\mathrm{diag} \{\exp(i\theta_1^{(s)}),\ldots, \exp(i\theta_d^{(s)})\}.
\end{aligned}
\end{equation}

Hence, Eq.~\ref{main_eq} is equivalent to
\begin{equation}
\begin{aligned}
\mathbf W_s\her \mathbf W_t &=\mathbf W_{t-s},\\
\mathbf P\her {\Lambda^{(s)}}\her \mathbf P \mathbf P\her \Lambda^{(t)}\mathbf P &= \mathbf P\her \Lambda^{(t-s)}\mathbf P,\\
\mathbf P\her {\mathbf \Lambda^{(s)}}\her\mathbf \Lambda^{(t)}\mathbf P &= \mathbf P\her\mathbf \Lambda^{(t-s)}\mathbf P,\\
{\mathbf \Lambda^{(s)}}\her\mathbf \Lambda^{(t)}&=\mathbf\Lambda^{(t-s)},\\
\mathrm{diag}\left\{
j(\theta_1^{(t)}-\theta_1^{(s)}),
j(\theta_2^{(t)}-\theta_2^{(s)}),\cdots, 
j(\theta_d^{(t)}-\theta_d^{(s)})\right\}&=
\mathrm{diag}
\left\{
j\theta_1^{(t-s)},
j\theta_2^{(t-s)},\cdots, 
j\theta_d^{(t-s)}\right\}.
\end{aligned}
\end{equation}
In this case, {$\forall k=1,\ldots, d$}, we have
\begin{equation}
\theta_k^{(t)}-\theta_k^{(s)} = \theta_k^{(t-s)} + 2l\pi, k,l\in \mathbb Z.
\end{equation}
Note that {$2l\pi$} does not affect the result, so we can assume {$l=0$}, \ig, 
\begin{equation}
\theta_k^{(t)}-\theta_k^{(s)} = \theta_k^{(t-s)}.
\end{equation}
Taking {$t=s+1$}, we get
\begin{equation}
\begin{aligned}
&\theta_k^{(s+1)} -\theta_k^{(s)}=\theta_k^{(1)},\\
&\theta_k^{(s)} =s \theta^{(1)}_k \triangleq s \alpha_k.
\end{aligned}
\end{equation}

\end{proof}

\subsection{Orthogonal (Solution 2)}
\begin{proof}[Proof of Proposition~\ref{p2}]
\label{proof4}
According to Theorem {\ref{th2}}, we can assume that $\mathbf W_s$ has the following form ({$\mathbf P\in \mathbb R^{d\times d}$} is an \textbf{orthogonal} matrix),
\begin{equation}
\begin{aligned}
\mathbf W_s &= \mathbf P\tran\mathbf \Lambda^{(s)}\mathbf P,\\
\mathbf\Lambda^{(s)} &= \left[\begin{matrix}
\mathbf A^{(s)} &  & \\ 
 &\mathbf B^{(s)} & \\
 & & \mathbf C^{(s)}
\end{matrix}\right],\\
\mathbf A^{(s)}&=\left[\begin{matrix}
\mathbf A_{1}^{(s)} &  \\ 
& \ddots   \\ 
& & \mathbf A_{n}^{(s)}  \\ 
\end{matrix}\right] \in \mathbb R^{2p\times 2p},\\
\mathbf B^{(s)}&= \mathbf I_q\in \mathbb R^{q\times q},\\
\mathbf C^{(s)}&= \mathbf -\mathbf I_r \in \mathbb R^{r\times r},\\
\mathbf A_k^{(s)}&=\left[\begin{matrix}
\cos \theta_{k}^{(s)} & -\sin \theta_{k}^{(s)} \\
\sin \theta_{k}^{(s)} & \cos \theta_{k}^{(s)}
\end{matrix}\right].
\end{aligned}
\end{equation}

Hence, Eq.~\ref{main_eq} is equivalent to
\begin{equation}
\begin{aligned}
\mathbf W_s\tran\mathbf W_t &=\mathbf W_{t-s},\\
\mathbf{P}\tran{\mathbf{\Lambda}^{(s)}}\tran\mathbf{P} \mathbf P\tran\mathbf \Lambda^{(t)}\mathbf P &= \mathbf{P}\tran\mathbf \Lambda^{(t-s)} \mathbf{P},\\
\mathbf P\tran {\mathbf \Lambda^{(s)}}\tran \mathbf\Lambda^{(t)}\mathbf P &= \mathbf P\tran\mathbf \Lambda^{(t-s)}\mathbf P,\\
{{\mathbf {\Lambda}}^{(s)}}\tran\mathbf \Lambda^{(t)}&=\mathbf \Lambda^{(t-s)},\\
\left[\begin{matrix}
 {\mathbf A^{(s)}}\tran & &    \\ 
 & {\mathbf B^{(s)}}\tran &    \\
  & &{\mathbf C^{(s)}}\tran   
\end{matrix}\right] 
\left[\begin{matrix}
\mathbf A^{(t)} & &    \\ 
 &\mathbf B^{(t)} &    \\
  & &\mathbf C^{(t)}   
\end{matrix}\right] &= 
\left[\begin{matrix}
\mathbf A^{(t-s)} & &    \\ 
 &\mathbf B^{(t-s)} &    \\
  & &\mathbf C^{(t-s)}   
\end{matrix}\right],
\end{aligned}
\end{equation}
where
\begin{equation}
\begin{aligned}
{\mathbf A^{(s)}}\tran \mathbf A^{(t)} &= \mathbf A^{(t-s)}, \\
{\mathbf B^{(s)}}\tran \mathbf B^{(t)} &= \mathbf B^{(t-s)}, \\
{\mathbf C^{(s)}}\tran \mathbf C^{(t)} &= \mathbf C^{(t-s)}. 
\end{aligned}
\end{equation}
For {$\mathbf A^{(s)}$}, considering the {$k$}-th component, we get
\begin{equation}
\begin{aligned}
{\mathbf A^{(s)}_k}\tran {\mathbf A}^{(t)}_k &= {\mathbf A}^{(t-s)}_k \\
&=\left[\begin{array}{cc}\cos  \theta^{(s)}_{k} & \sin \theta^{(s)}_{k} \\ -\sin \theta^{(s)}_{k} & \cos \theta^{(s)}_{k} \end{array}\right]
\left[\begin{array}{cc}\cos  \theta^{(t)}_{k} & -\sin \theta^{(t)}_{k} \\ \sin \theta^{(t)}_{k} & \cos \theta^{(t)}_{k} \end{array}\right]\\
&=\left[\begin{array}{cc}
\cos \theta^{(s)}_{k} \cos \theta^{(t)}_{k} +\sin \theta^{(s)}_{k}\cos \theta^{(t)}_{k}  &
\sin \theta^{(s)}_{k}  \cos \theta^{(t)}_{k} - \cos \theta^{(s)}_{k}  \sin\theta^{(t)}_{k} \\ 
-\sin  \theta^{(s)}_{k} \cos\theta^{(t)}_{k}   +\cos\theta^{(s)}_{k} \sin \theta^{(t)}_{k} 
& \cos\theta^{(s)}_{k} \cos \theta^{(t)}_{k}  +\sin\theta^{(s)}_{k} \sin \theta^{(t)}_{k}  
\end{array}\right]\\
&=\left[\begin{array}{cc}
\cos \left(\theta^{(t)}_{k}-\theta^{(s)}_{k} \right)  &
-\sin \left(\theta^{(t)}_{k} -\theta^{(s)}_{k} \right) \\ 
\sin \left(\theta^{(t)}_{k} -\theta^{(s)}_{k} \right)
&\cos \left(\theta^{(t)}_{k} -\theta^{(s)}_{k} \right) \\
\end{array}\right]\\
&=\mathbf A^{(t-s)}_{k} \\
&=\left[\begin{array}{cc}\cos  \theta^{(t-s)}_{k} & -\sin \theta^{(t-s)}_{k} \\ \sin \theta^{(t-s)}_{k} & \cos \theta^{(t-s)}_{k} \end{array}\right].
\end{aligned}
\end{equation}
Hence, {$\forall k=1,\ldots, d$}, we have
\begin{equation}
\theta_k^{(t)}-\theta_k^{(s)} = \theta_k^{(t-s)} + 2k\pi, k\in \mathbb Z.
\end{equation}
Note that {$2t\pi$} does not affect the result, so  we can assume {$t=0$}, \ig, 
\begin{equation}
\theta_k^{(t)}-\theta_k^{(s)} = \theta_k^{(t-s)}.
\end{equation}
Taking {$t=s+1$}, we have
\begin{equation}
\begin{aligned}
&\theta_k^{(s+1)} -\theta_k^{(s)}=\theta_k^{(1)},\\
&\theta_k^{(s)} =s \theta^{(1)}_k \triangleq s \alpha_k.
\end{aligned}
\end{equation}
Next, for {$\mathbf B^{(s)}$}, the conclusion is more obvious, \ig,
\begin{equation}
\begin{aligned}
{\mathbf B^{(s)}}\tran \mathbf B^{(t)}
&= \mathbf I_{q}\tran \mathbf I_q \\
&=\mathbf I_q\\
&= \mathbf B^{(t-s)}. \\
\end{aligned}
\end{equation}
Finally, for {$\mathbf C^{(s)}$}, we have
\begin{equation}
\begin{aligned}
{\mathbf C^{(s)}}\tran \mathbf C^{(t)}
&= (-\mathbf I_{r}^T )(-\mathbf I_r) \\
&=\mathbf I_r\\
&\neq \mathbf C^{(t-s)}. \\
\end{aligned}
\end{equation}
In that case, we must have {$r=0$}.

\end{proof}

\subsection{Permutation}
\label{proof5}
Prior to the proof, we first provide some relevant definitions and propositions.

\newtheorem{de_permutate}[de_lrpe]{Definition}

\begin{de_permutate}
Permutation {$\pi$} is a \textbf{bijection} defined on the integer set:
\begin{equation}
\pi:\{1,2, \cdots, d\} \rightarrow \{1,2, \cdots, d\}, d\in \mathbb Z^+.
\end{equation}
\end{de_permutate}

\newtheorem{de_peronmatrix}[de_lrpe]{Definition}

\begin{de_permutate}
For matrix 
\begin{equation}
\mathbf M=\left[\begin{matrix}
\mathbf m_1\tran \\
\mathbf m_2\tran \\
\vdots\\
\mathbf m_d\tran
\end{matrix}\right] \in \mathbb R^{d\times d},\mathbf m_k \in \mathbb R^d , k=1,\ldots ,d,
\end{equation}
{$\mathbf M_{\pi}$} is defined as
\begin{equation}
\mathbf M_\pi=\left[\begin{matrix}
\mathbf m_{\pi(1)}\tran \\
\mathbf m_{\pi(2)}\tran \\
\vdots\\
\mathbf m_{\pi(d)}\tran
\end{matrix}\right]. 
\end{equation}

\end{de_permutate}

\newtheorem{de_perid}[de_lrpe]{Definition}

\begin{de_perid}
\label{de_perid}
For identity matrix {$\mathbf I_d\in \mathbb R^{d\times d}$} and permutation {$\pi$}, we define
\begin{equation}
\mathbf \Lambda_k = (\mathbf I_d)_{\pi^k}.
\end{equation}

\end{de_perid}

For {$\mathbf \Lambda_k$}, we have the following important properties:

\newtheorem{le_per}[de_lrpe]{Lemma}

\begin{le_per}
\label{le_per}
For permutation $\pi$, matrix {$\mathbf M\in \mathbb R^{d\times d}$} and {$\mathbf \Lambda_k \in \mathbb R^{d \times d}$} defined in \ref{de_perid}, 
we have
\begin{equation}
\mathbf M_{\pi} = \mathbf \Lambda_1 \mathbf M.
\end{equation}
\begin{proof}
We first organize {$\mathbf I_d\in \mathbb R^{d\times d}$} in the following form, where {$\mathbf e_k\in \mathbb R^{d},k=1,\ldots,d$} represents the one-hot vector with the {$k$}-th element as one, \ig, 
\begin{equation}
\mathbf I_d=\left[\begin{array}{c}{\mathbf e}_{1}\tran \\ {\mathbf e}_{2}\tran \\ \vdots \\ {\mathbf e}_{d}\tran\end{array}\right].
\end{equation}
Notice that 
\begin{equation}
\mathbf e_k\tran \mathbf M = \mathbf{m}_{k}\tran,
\end{equation}
so we get
\begin{equation}
\begin{aligned}
\mathbf \Lambda_1 \mathbf M
&= \left[\begin{array}{c}{\mathbf e}_{\pi(1)}\tran \\ {\mathbf e}_{\pi(2)}\tran \\ \vdots \\ {\mathbf e}_{\pi(d)}\tran\end{array}\right]\mathbf M\\
&= \left[\begin{array}{c}{\mathbf e}_{\pi(1)}\tran\mathbf  M \\ {\mathbf e}_{\pi(2)}\tran\mathbf  M \\ \vdots \\ {\mathbf e}_{\pi(d)}\tran\mathbf  M\end{array}\right] \\
&= \left[\begin{array}{c}\mathbf m_{\pi(1)}\tran \\ \mathbf m_{\pi(2)}\tran \\ \vdots \\ \mathbf m_{\pi(d)}\tran\end{array}\right]\\
&= \mathbf M_{\pi}.
\end{aligned}
\end{equation}

\end{proof}
\end{le_per}

\newtheorem{th_per}[de_lrpe]{Theorem}
\begin{th_per}
\label{th3}
For $\mathbf \Lambda_k$ defined in \ref{de_perid}, we have:
\begin{equation}
\mathbf \Lambda_k= \mathbf \Lambda_1^k.
\end{equation}

\begin{proof}

We use induction for the proof.

For {$k=1$}, the conclusion is obvious. Now assuming that the conclusion holds for {$k=s-1$}, when $k=s$, we have
\begin{equation}
\begin{aligned}
\mathbf \Lambda_{s}
&=(\mathbf I_d)_{\pi^s}\\
&=((\mathbf I_d)_{\pi^{s-1}})_{\pi}\\
&=(\mathbf \Lambda_{s-1})_{\pi}\\
&= (\mathbf \Lambda_1^{s-1})_{\pi}.
\end{aligned}
\end{equation}
The next step is to prove
\begin{equation}
\begin{aligned}
(\mathbf \Lambda_1^{s-1})_{\pi}= \mathbf \Lambda_1^s =\mathbf \Lambda_1 \mathbf \Lambda_1^{s-1}. 
\end{aligned}
\end{equation}
The above conclusion follows from \ref{le_per}.

\end{proof}
\end{th_per}

\newtheorem{th_per_or}[de_lrpe]{Theorem}

\begin{th_per_or}
\label{th4}
{$\mathbf \Lambda_k\in \mathbb R^{d \times d}$} defined in \ref{de_perid} are orthogonal matrices, i.e.,
\begin{equation}
\mathbf \Lambda_k \mathbf \Lambda_k^T= \mathbf \Lambda_k\tran\mathbf \Lambda_k=\mathbf I_d.
\end{equation}

\begin{proof}
We first prove that the conclusion holds for {$k=1$}:
\begin{equation}
\begin{aligned}
\mathbf \Lambda_1 \mathbf \Lambda_1\tran
&= \left[\begin{array}{c}{\mathbf e}_{\pi(1)}\tran \\ {\mathbf e}_{\pi(2)}\tran \\ \vdots \\ {\mathbf e}_{\pi(d)}\tran\end{array}\right]
\left[\begin{matrix}{\mathbf e}_{\pi(1)} & {\mathbf e}_{\pi(2)} & \ldots & {\mathbf e}_{\pi(d)}\end {matrix}\right],\\
\left[\mathbf \Lambda_1 \mathbf \Lambda_1\tran\right]_{st}
&= {\mathbf e}_{\pi(s)}\tran{\mathbf e}_{\pi(t)} \\
&=\delta_{st}, \\
\mathbf \Lambda_1 \mathbf \Lambda_1\tran & = \mathbf I_d.
\end{aligned}
\end{equation}
Since {$\mathbf \Lambda_1$} is a square matrix, we also have
\begin{equation}
\mathbf \Lambda_1\tran\mathbf \Lambda_1=\mathbf I_d.
\end{equation}
In general cases, we only use \ref{th3}, \ig, 
\begin{equation}
\begin{aligned}
\mathbf \Lambda_k \mathbf \Lambda_k\tran
&= \mathbf \Lambda_1^k  ( \mathbf \Lambda_1^k)\tran \\
&= \mathbf \Lambda_1^k(\mathbf \Lambda_1\tran)^k \\
&= \mathbf \Lambda_1^{k-1}\mathbf \Lambda_1 \mathbf \Lambda_1\tran (\mathbf \Lambda_1\tran)^{k-1} \\
&= \mathbf \Lambda_1^{k-1}(\mathbf \Lambda_1\tran)^{k-1}\\
&= \ldots \\
&= \mathbf I_d.
\end{aligned}
\end{equation}
With the same proof, we get
\begin{equation}
\mathbf \Lambda_k\tran\mathbf \Lambda_k=\mathbf I_d.
\end{equation}
\end{proof}
\end{th_per_or}

Based on the above conclusions, we can prove Proposition \ref{p3} below.

\begin{proof}[Proof of Proposition~\ref{p3}]
According to Theorem {\ref{th4}} and the product of the \textbf{orthogonal} matrix is an \textbf{orthogonal} matrix, we can assume that $\mathbf W_k$ has the following form ({$\mathbf P\in \mathbb R^{d\times d}$} is an \textbf{orthogonal} matrix), \ig,
\begin{equation}
\begin{aligned}
\mathbf W_k &= \mathbf P\tran \mathbf \Lambda^{(k)} \mathbf P.\\
\end{aligned}
\end{equation}
The next step is to verify that it satisfies Eq.~\ref{main_eq}, which follows Theorem {\ref{th3}} and {\ref{th4}}:
\begin{equation}
\label{eq:trick}
\begin{aligned}
\mathbf W_s\tran \mathbf W_t
&=  \mathbf P\tran {\mathbf \Lambda^{(s)}}\tran\mathbf P \mathbf P\tran {\mathbf \Lambda^{(t)}} \mathbf P \\
&=  \mathbf P\tran {\mathbf \Lambda^{(s)}}\tran {\mathbf \Lambda^{(t)}} \mathbf P \\
&=  \mathbf P\tran {\mathbf \Lambda^{(s)}}\tran (\mathbf \Lambda^{(1)})^t \mathbf P \\
&=  \mathbf P\tran {\mathbf \Lambda^{(s)}}\tran (\mathbf \Lambda^{(1)})^s (\mathbf \Lambda^{(1)})^{t-s} \mathbf P \\
&=  \mathbf P\tran {\mathbf \Lambda^{(s)}}\tran {\mathbf \Lambda^{(s)}} (\mathbf \Lambda^{(1)})^{t-s} \mathbf P \\
&=  \mathbf P\tran {\mathbf \Lambda^{(t-s)}} \mathbf P \\
&=  \mathbf W_{t-s}.
\end{aligned}
\end{equation}

\end{proof}

% \end{p3}

\section{Implementation}
\subsection{Theory}
\label{imp_detail}
{\name}($\mathbf W_s =\mathbf P\her \mathbf\Lambda^{(s)}\mathbf P$) contains two components, \ig, the fixed unitary matrix $\mathbf P$ and the unitary matrix family $\mathbf \Lambda^{(s)}$ mentioned in proposition \ref{p1}, \ref{p2}, and \ref{p3}. 
We first introduce the choice of matrices $\mathbf P / {\mathbf \Lambda}^{(s)}$, and then illustrate some implementation tricks.

\paragraph{Choice of matrices}\mbox{} 

For matrix $\mathbf P$, We list the species mentioned in the paper below:
\begin{itemize}
    \item Householder matrix: denoted as a vector $\mathbf v\in \mathbb R^d$, \ig, 
\begin{equation}
\begin{aligned}
\mathbf W =\mathbf  I_d - 2 \mathbf v\mathbf  v\tran / (\mathbf v\tran \mathbf v).
\end{aligned}
\end{equation}
In our implementation, we sample $\mathbf v$ from a standard normal distribution, and make it \textbf{deterministic}.
    \item Permutation matrix: formulated as per the following permutation (inspired by Flash \citep{hua2022transformer}), \ig, 
\begin{equation}
\begin{aligned}
\pi (2k) = k, \pi(2k+1) = \lfloor d/ 2\rfloor + 1, 1\le 2k, 2k+1 \le d.
\end{aligned}
\end{equation}
\item Identity matrix.
\end{itemize}

% (1) identity matrix, (2) householder matrix \citep{golub2013matrix}, (3) permutation matrix, and (4) FFT matrix.
% The identity matrix is trivial and needs no explanation. 
% The Householder matrix is described by a vector v∈Rd\mathbf v\in \mathbb R^d:
% \begin{equation}
% \begin{aligned}
% \mathbf W =\mathbf  I_d - 2 \mathbf v\mathbf  v\tran / (\mathbf v\tran \mathbf v).
% \end{aligned}
% \end{equation}
% In our implementation, we sample v\mathbf v from standard normal distribution, and make it \textbf{deterministic} or \textbf{learnable}.

% The permution matrix corresponds to the following permution, which is inspired by Flash \citep{hua2022transformer}:
% \begin{equation}
% \begin{aligned}
% \pi (2k) = k, \pi(2k+1) = \lfloor d/ 2\rfloor + 1, 1\le 2k, 2k+1 \le d.
% \end{aligned}
% \end{equation}
% FFT matrix is the matrix form of FFT (Fast Fourier Transform).

For \textbf{matrix family} $\mathbf \Lambda^{(s)}$, we use the following settings:
\begin{itemize}
\item For unitary (Solution 1) (\ref{p1}), we use the same method in \citep{su2021roformer} with initialized $\alpha_t=10000^{-2t /d}$, and make it learnable.
\item For orthogonal (Solution 2) (\ref{p2}), we choose the dimension of identity submatrix $q=0$ with initialized $\alpha_t=10000^{-2t /d}$ as in \citep{su2021roformer} and make it \textbf{learnable}.
\begin{itemize}
    \item Another notable version to choose the dimension of the identity submatrix $q=0$ with initialized $\alpha_t=10000^{-2t /d}$ as in \citep{su2021roformer}, and make it \textbf{deterministic}. When using this version along with the identity matrix, we can get \textbf{RoPE} \citep{su2021roformer}.
\end{itemize}
\item For permutation (Solution 3) (\ref{p3}), we randomly choose the permutation and make it \textbf{deterministic}.
\begin{itemize}
\item Notice that when combing this method with identity matrix, we can get a version of \textbf{PermuteFormer} \citep{chen2021permuteformer}.
\end{itemize}
\end{itemize}

\paragraph{Implementation tricks}\mbox{}

According to the following facts, we can simplify the computation, \ig,
\begin{equation}
\begin{aligned}
\mathbf q_s\her\mathbf  W_s\her\mathbf  W_t\mathbf  k_t 
&= \mathbf q_s\her\mathbf P\her (\mathbf \Lambda^{(s)})^{\her}\mathbf P 
\mathbf P\her  \mathbf \Lambda^{(t)}\mathbf P \mathbf  k_t\\
&=  \mathbf q_s\her\mathbf P\her (\mathbf \Lambda^{(s)})^{\her} \mathbf \Lambda^{(t)}\mathbf P \mathbf  k_t\\
&= (\mathbf \Lambda^{(s)}\mathbf P \mathbf q_s)\her (\mathbf \Lambda^{(t)}\mathbf P \mathbf  k_t).
\end{aligned}
\end{equation}
Hence, in practice, we can use $\mathbf W_s =\mathbf P\her \mathbf\Lambda^{(s)}$ instead of $\mathbf W_s =\mathbf P\her \mathbf\Lambda^{(s)}\mathbf P$ to reduce the computational costs.

\subsection{Pseudocode}
In this section, we provide pseudocodes for {\name} in Python:
\begin{lstlisting}
import torch
import torch.nn as nn
import numpy as np

class Lrpe(nn.Module):
    def __init__(self, core_matrix, p_matrix, max_positions=512, embedding_dim=768, 
                 theta_type="a", theta_learned=False, householder_learned=False):
        super().__init__()
        self.core_matrix = core_matrix
        self.p_matrix = p_matrix
        self.theta_type = theta_type
        self.theta_learned = theta_learned
        self.householder_learned = householder_learned

        # Lambda matrix
        if self.core_matrix == 1:
            if self.theta_learned:
                print("Learn theta!")
                self.theta = nn.Parameter(10000 ** (-2 / embedding_dim * torch.arange(embedding_dim // 2)).reshape(1, 1, -1))
            else:
                print(f"Theta_type {self.theta_type}")
        elif self.core_matrix == 2:
            print("Mixed")
        elif self.core_matrix == 3:
            print("Permutation")
            permutation = self.get_permutation(max_positions, embedding_dim)
            self.register_buffer("permutation", permutation)
        elif self.core_matrix == 4:
            print("Complex exp")
            if self.theta_learned:
                print("Learn theta!")
                self.theta = nn.Parameter(10000 ** (-2 / embedding_dim * torch.arange(embedding_dim)).reshape(1, 1, -1))
            else:
                print(f"Theta_type {self.theta_type}")

        # P matrix
        if self.p_matrix == 1:
            print("Identity")
        elif self.p_matrix == 2:
            print("Householder")
            if self.householder_learned:
                print("learn householder!")
                self.v = nn.Parameter(torch.randn(1, embedding_dim, 1))
            else:
                v = torch.randn(1, embedding_dim, 1)
                v = v / torch.norm(v)
                print(f"Householder norm is {torch.norm(v)}")
                self.v = nn.Parameter(v, requires_grad=False)
        elif self.p_matrix == 3:
            print("Fourier")
        elif self.p_matrix == 4:
            print("Odd_even")

        self.p = self.get_p()
        self.core_transform = self.get_core_transform()

    def forward(self, x):
        '''
        input shape: (b, l, e), b stands for batch size, l stands for sequence length, e stands for embedding dimension.
        '''
        x = self.p(x)
        x = self.core_transform(x)
        return x

    def get_p(self):
        if self.p_matrix == 1:
            def f(x):
                return x
            return f
        elif self.p_matrix == 2:
            return self.householder
        elif self.p_matrix == 3:
            def f(x):
                return torch.fft.fft(x, norm="ortho")
            return f
        elif self.p_matrix == 4:
            return self.odd_even_permutation

    def get_core_transform(self):
        if self.core_matrix == 1:
            return self.reflect
        elif self.core_matrix == 2:
            return self.mix_reflect
        elif self.core_matrix == 3:
            return self.do_permutation
        elif self.core_matrix == 4:
            return self.complex_exp

    def get_permutation(self, max_positions, embedding_dim):
        permutation = torch.randperm(embedding_dim).reshape(1, -1)
        expanded = [torch.arange(embedding_dim).unsqueeze(0)]
        for _ in range(max_positions - 1):
            previous = expanded[-1]
            current = previous.gather(-1, permutation)
            expanded.append(current)
        expanded = torch.stack(expanded, dim=1)
        return expanded

    def odd_even_permutation(self, x):
        # 2k->k, 2k+1->d+k
        e = x.shape[-1]
        d = e - e // 2
        permutation = torch.arange(e)
        index = torch.arange(e)
        permutation[::2] = index[::2] // 2
        permutation[1::2] = (index[1::2] - 1) // 2 + d
        permutation = permutation.to(x.device)
        x = x.gather(-1, permutation.expand_as(x))

        return x

    def do_permutation(self, x):
        b, l, e = x.shape
        x = x.gather(-1, self.permutation[:, :l, :].expand_as(x))

        return x

    def reflect(self, x):
        b, l, d = x.shape
        e = d - 1 if d % 2 == 1 else d
        return self.transform(x, e)

    def mix_reflect(self, x):
        b, l, d = x.shape
        assert d >= 3
        # split
        e = d // 2
        # to even
        if e % 2:
            e += 1
        return self.transform(x, e)

    def transform(self, x, e):
        assert e % 2 == 0
        b, l, d = x.shape
        # do identity transformation
        x1 = x[:, :, e:]
        # do reflection
        x = x[:, :, :e]
        if self.theta_learned:
            theta = self.theta
        else:
            if self.theta_type == "a":
                theta = 10000 ** (-2 / e * torch.arange(e // 2))
            elif self.theta_type == "b":
                theta = np.pi / 2 / l / (e // 2) * torch.arange(1, e // 2 + 1)
            elif self.theta_type == "c":
                theta = np.pi / 2 / l / torch.arange(1, e // 2 + 1)
            theta = theta.reshape(1, 1, -1).to(x)
        theta = torch.stack([theta, theta], dim=-1).reshape(1, 1, e)
        theta = theta * torch.arange(l).reshape(1, -1, 1).to(x)
        # (-q1, -q3), (q0, q2) -> (-q1, q0, -q3, q2)
        x_half = torch.stack([-x[..., 1::2], x[..., ::2]], dim=-1).reshape_as(x)
        x_transform = x * torch.cos(theta) + x_half * torch.sin(theta)
        # merge
        if e != d:
            x_transform = torch.cat([x_transform, x1], dim=-1)

        return x_transform

    def complex_exp(self, x):
        b, l, e = x.shape
        if self.theta_learned:
            theta = self.theta
        else:
            if self.theta_type == "a":
                theta = 10000 ** (-2 / e * torch.arange(e))
            theta = theta.reshape(1, 1, -1).to(x.device)
        matrix = theta * torch.arange(l).reshape(1, -1, 1).to(x.device)

        sin_cos = torch.complex(torch.cos(matrix),torch.sin(matrix)).to(x.device)
        x = self.element_wise_complex(x, sin_cos)
        return x

    def element_wise_complex(self, t1, t2):
        return torch.complex(t1.real * t2.real - t1.imag * t2.imag, t1.real * t2.imag + t1.imag * t2.real)

    def householder(self, x, eps=1e-6):
        if self.householder_learned:
            v = self.v / (torch.norm(self.v) + eps)
        else:
            v = self.v
        # (b, n, e), (1, e, 1) -> (1, n, 1)
        y = torch.matmul(x, v)
        # (1, n, 1), (1, 1, e) -> (1, n, e)
        y = torch.matmul(y, v.transpose(1, 2))

        return x - 2 * y
\end{lstlisting}

\section{Configuration}
\label{config}

\begin{table*}[!ht]
\center
\small
\setlength{\tabcolsep}{3.4mm}{
\caption{Detailed configurations used in our experiments. ``Total batch size'' means $\mathrm{batch\_per\_gpu} \times \mathrm{update\_freq} \times \mathrm{num\_gpus}$. ``Attention dropout'' is only used for vanilla attention. ``ALM'': autoregressive Language Model. ``BLM'': bidirectional Language Model. ``IM'': Image Modeling. }
\label{configuration}
\begin{tabular}{l|l|l|l}
\hline\hline
  & ALM & BLM & IM \\
\hline\hline
Data    & WikiText-103/Books  & Wikibook    & ImageNet-1k          \\
Tokenizer method & BPE    & BPE & -\\
Vocab size & 267744/50265  & 50265 &  - \\
Encoder layers        & 0         & 12     & 12                     \\
Decoder layers           & 6          & 0   & 0                     \\
Hidden dimensions       & 512       & 768          & 384                   \\
Number of heads           & 8      & 12    & 6        \\
FFN dimensions       & 2048       & 3072       & 1536  \\
FFN activation function     & Relu    & Gelu      & Gelu                 \\
Seqence length        & 512        & 51      & -   \\
Total batch size & 128    & 512       & 1600                   \\
Number of updates    & 50k updates   & 23k updates      & 300 epochs             \\
Warmup steps    & 4k steps    & 3k steps    & 20 epochs                    \\
Peak learning rate        & 5e-4                      & 5e-4                    & 5e-4             \\
Learning rate scheduler    & Inverse sqrt  & Polynomial decay             & Cosine          \\
Optimizer    & Adam  & Adam         & Adamw                 \\
Adam $\epsilon$           & 1e-8                      & 1e-6                     & 1e-8              \\
Adam $(\beta_1,\beta_2)$                                                   & (0.9, 0.98)                   & (0.9, 0.98)                  & (0.9, 0.98)           \\
Weight decay          & 0.01         & 0.01       & 0.05    \\              
\hline\hline
\end{tabular}}

\end{table*}

\begin{table}[!ht]
\caption{Detailed configurations used in LRA experiments. `BN` stands for batch normalization. All methods use the same configuration, except for relative positional encodings. }
    \centering
    \begin{tabular}{l|l|l|l|l|l|l}
    \hline
    \hline
        Task & Feature dim & Layer & Norm & Batch size & Epoch & Lr \\ \hline \hline
        Text & 128 & 4 & BN & 256 & 32 & 0.001 \\ 
         ListOps  & 128 & 4 & BN & 256 & 40 & 0.0001 \\ 
        Retrieval  & 64 & 4 &  BN & 64 & 20 & 0.001 \\
        Pathfinder  & 32 & 4 & BN & 128 & 200 & 0.0005 \\ 
        Image & 100 & 12 &  BN & 100 & 200 & 0.001 \\ \hline \hline
    \end{tabular}
\end{table}

\end{document}